\documentclass[10pt,twocolumn,letterpaper]{article}
\usepackage[pagenumbers]{style/iccv}

\newcommand{\rcnt}[1]{\rotatebox[origin=c]{90}{#1}}
\newcommand{\xhdr}[1]{\vspace{4pt}\noindent\textbf{#1}}

\definecolor{myblue}{HTML}{4E79A7}
\definecolor{myorange}{HTML}{F28E2B}
\definecolor{mygreen}{HTML}{59A14F}
\definecolor{mypurple}{HTML}{B07AA1}
\definecolor{myred}{HTML}{E15759}
\definecolor{mygray}{HTML}{4C535D}

\colorlet{standard}{myblue}
\colorlet{best}{myorange}
\colorlet{average}{mygreen}
\colorlet{bpe}{mypurple}
\colorlet{mmb}{myred}

\newcommand{\myquote}[1]{``\emph{#1}''}

\newcommand{\std}{\textsc{Std.}\xspace}
\newcommand{\bst}{\textsc{Best}\xspace}
\newcommand{\avg}{\textsc{Avg.}\xspace}
\newcommand{\bpe}{\textsc{BPE}\xspace}
\newcommand{\mmb}{\textsc{MMB}\xspace}
\newcommand{\mmblong}{\textbf{M}ultimodal \textbf{M}ixture-of-\textbf{B}ayesian Prompt Ensembles\xspace}
\newcommand{\mycirc}[1]{\Circled[fill color=mygray, inner color=white, outer color=mygray]{#1}}
\usepackage{paralist}
\usepackage[dvipsnames]{xcolor}
\usepackage{booktabs}
\usepackage{multirow}
\usepackage{enumitem}
\usepackage{circledsteps}
\usepackage{bbding}
\usepackage{makecell}
\usepackage{tcolorbox}
\usepackage{colortbl}
\usepackage{cuted}
\usepackage{listings}
\definecolor{codegreen}{rgb}{0,0.6,0}
\definecolor{codegray}{rgb}{0.5,0.5,0.5}
\definecolor{codepurple}{rgb}{0.58,0,0.82}
\definecolor{backcolour}{rgb}{0.95,0.95,0.95}

\lstdefinestyle{mystyle}{
    backgroundcolor=\color{backcolour},   
    commentstyle=\color{codegreen},
    keywordstyle=\color{codepurple},
    numberstyle=\tiny\color{codegray},
    stringstyle=\color{codepurple},
    basicstyle=\ttfamily\footnotesize,
    breakatwhitespace=false,         
    breaklines=true,                 
    captionpos=b,                    
    keepspaces=true,                 
    numbers=left,                    
    numbersep=5pt,                  
    showspaces=false,                
    showstringspaces=false,
    showtabs=false,                  
    tabsize=2,
}

\lstdefinelanguage{PyTorch}{%
  language     = Python,
  morekeywords = {ones, T, where, argmax, update, shape, get_min_concept, log_and_keep},
}

\setitemize{noitemsep,topsep=0pt,parsep=0pt,partopsep=0pt}

\definecolor{iccvblue}{rgb}{0.21,0.49,0.74}
\usepackage[pagebackref,breaklinks,colorlinks,allcolors=iccvblue]{hyperref}
\usepackage{fvextra}

\def\paperID{14610}
\def\confName{ICCV}
\def\confYear{2025}

\title{Calibrating MLLM-as-a-judge via Multimodal Bayesian Prompt Ensembles\vspace{-0.5em}}
\author{Eric Slyman$^{1,2}$\quad Mehrab Tanjim$^{1}$ \quad Kushal Kafle$^{1}$\quad Stefan Lee$^{2}$ \vspace{0.3em} \\
{\normalsize $^1$Adobe} \quad
{\normalsize $^2$Oregon State University} 
}

\begin{document}
\maketitle

\begin{abstract}
Multimodal large language models (MLLMs) are increasingly used to evaluate text-to-image (TTI) generation systems, providing automated judgments based on visual and textual context. However, these ``judge'' models often suffer from biases, overconfidence, and inconsistent performance across diverse image domains. While prompt ensembling has shown promise for mitigating these issues in unimodal, text-only settings, our experiments reveal that standard ensembling methods fail to generalize effectively for TTI tasks. To address these limitations, we propose a new multimodal-aware method called \mmblong (MMB). Our method uses a Bayesian prompt ensemble approach augmented by image clustering, allowing the judge to dynamically assign prompt weights based on the visual characteristics of each sample. We show that MMB improves accuracy in pairwise preference judgments and greatly enhances calibration, making it easier to gauge the judge’s true uncertainty. In evaluations on two TTI benchmarks, HPSv2 and MJBench, MMB outperforms existing baselines in alignment with human annotations and calibration across varied image content. Our findings highlight the importance of multimodal-specific strategies for judge calibration and suggest a promising path forward for reliable large-scale TTI evaluation.
\end{abstract}    

\vspace{-1em}
\section{Introduction} \label{sec:introduction}

Modern large vision-language models (LVLMs) and multimodal large language models (MLLMs) \citep{openai2023gpt4v, awadalla2023openflamingo,peng2023kosmos,bai2023qwen,liu2023llava,zhang2024llama, dai2024instructblip,zhu2024minigpt} are rapidly advancing in their ability to understand and respond to human-like instructions across various tasks. In general, these models can interpret both textual and visual content through a unified natural language interface, such as image-captioning~\citep{lin2014coco,young2014flickr}, visual question answering~\citep{antol2015vqa}, visual dialogue~\citep{das2017visdial}, and more~\citep{li2023seed}. Similarly, text-to-image (TTI) generators \citep{rombach2022high,podell2023sdxl,betker2023improving,cho2023dalle} can invert this process and render new images from textual prompts. Recent efforts have even begun to consolidate diverse multimodal capabilities into low-technical barrier unified ecosystems such as OpenAI-o1~\cite{openai2025o1} with DALL-E~\cite{cho2023dalle}, or Gemini~\cite{team2023gemini} with Imagen~\cite{saharia2022photorealistic}) which can do both analyses \eg \myquote{Please caption this photo} and generation \eg \myquote{Draw me an image of a space explorer!}

\begin{figure}[!t]
    \centering
    \includegraphics[width=\linewidth]{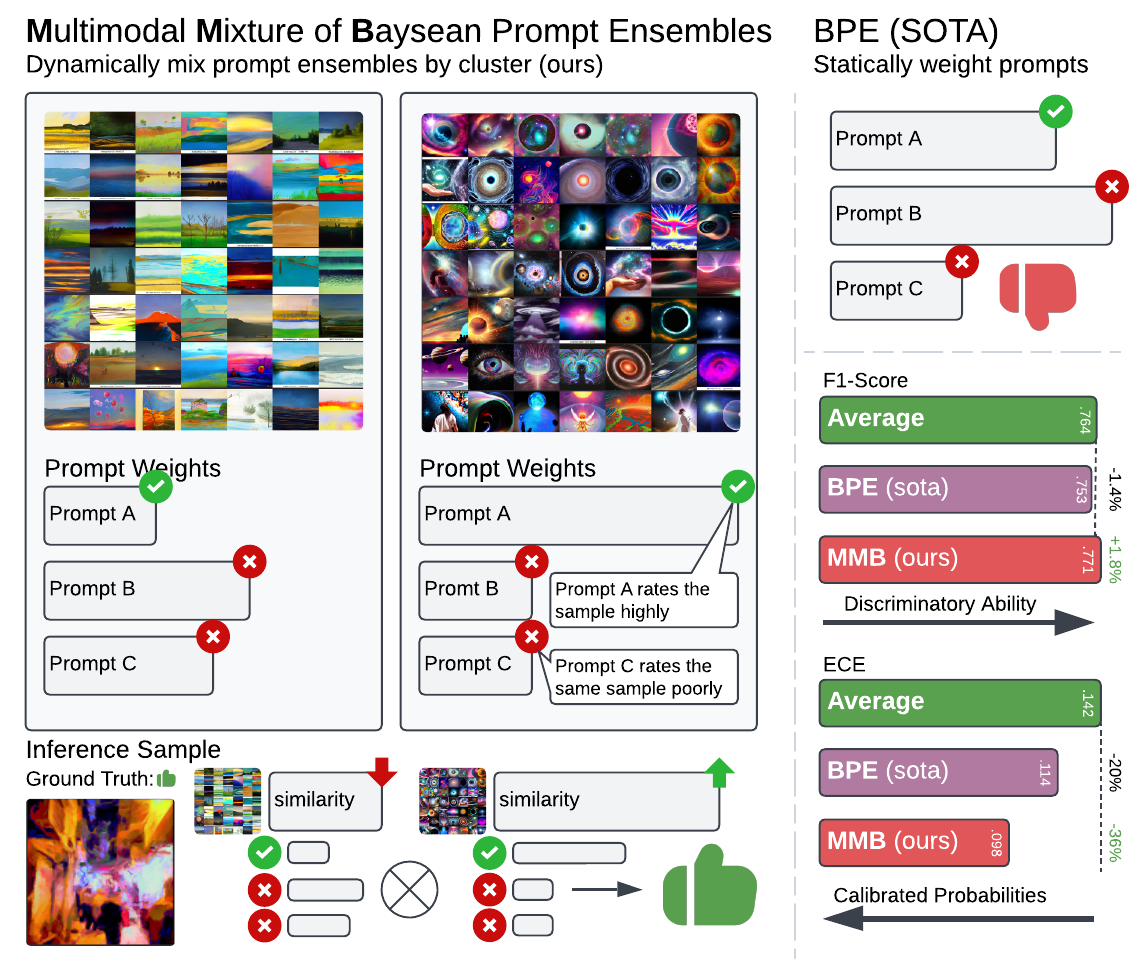}
    \vspace{-0.5em}
    \caption{A high-level comparison on the HPSv2~\citep{wu2023human} dataset between an average ensemble baseline (\avg), current SOTA \bpe~\citep{tonolini2024bayesian}, and our \mmb method. We show F1 Score (higher is better $\uparrow$) and ECE (lower is better $\downarrow$). While BPE somewhat lowers the model’s discriminative ability (F1) relative to the baseline, MMB both recovers that loss (+1.8\% F1 vs.\ \avg) and achieves stronger calibration (-36\% ECE vs.\ \avg). All differences are statistically significant at 95\% confidence via a permutation test.}
    \label{fig:teaser-orig}
\end{figure}

As a result, these models produce diverse and frequently subjective multimodal outputs, which makes evaluating them a significant challenge.
Traditional metrics for text (BLEU~\cite{papineni2002bleu}, ROUGE~\citep{lin2004rouge}, SPICE~\citep{anderson2016spice}) and image generation quality (\eg, FID~\cite{heusel2017gans}, Inception~\cite{salimans2016improved}, Precision-Recall~\cite{kynkaanniemi2019improved}) often fail to capture the open-ended, creative nature of responses that generative models can produce~\citep{zheng2023judging}. TTI generation, in particular, must be judged on aesthetic qualities, coherence with textual prompts, realism, and creativity dimensions that are often subjective and difficult to evaluate using fixed metrics. A challenge that has led to an array of preference scores, including CLIPScore~\citep{radford2021learning}, PickScore~\citep{kirstain2023pick}, and PreferenceScore~\cite{wu2023human}, that attempt to generally capture the \myquote{goodness} of an image. On the other hand, human evaluation—although more reliable for subtle qualities like image realism or appropriateness—can be too costly or slow to be practical at scale.

A related trend in the unimodal (text-only) domain involves using large language models (LLMs) themselves as judges to evaluate text-generation quality~\citep{li2024llmasajudge,li2024llms,gu2024survey}. This concept is also readily extended to LVLMs and MLLMs being used as judges
through frameworks like GPT-4V(ision)~\citep{zhang2023gpt,cui2024exploring}, X-IQE~\citep{chen2023x}, MLLM-Bench~\citep{ge2023mllm}, ViGOR~\citep{yan2024vigor}, and even Text-to-3D~\citep{wu2023gpteval3d}. Yet while these \myquote{judge} models can approximate human assessments of relevance, clarity, and creativity, they still exhibit biases. For instance, they may favor outputs from sharing their training lineage~\citep{liu2023llms,panickssery2024llm}, reward verbosity~\citep{saito2023verbosity}, or change their evaluations when prompts are slightly altered~\citep{wang2023large}. They often struggle with commonsense reasoning~\citep{he2022blind,koo2023benchmarking,hu2024are} and can inadvertently amplify social biases~\citep{ye2024justice,li2024llmasajudge}. Addressing these pitfalls is crucial to ensuring fair and accurate assessments of advanced multimodal model capabilities.

A further complication is that many of the most performant and accessible MLLMs are closed-source and available only through restricted APIs. This limitation curtails researchers’ ability to fine-tune or directly inspect the judge for a given use case. Recent work proposes ensembling and reweighting prompts~\citep{wightman2023strength, jiang2023calibrating, hou2023promptboosting, tonolini2024bayesian} to partially mitigate these issues by improving calibration and accuracy in black-box large language models. At a high level, a calibrated model's probabilities align with actual outcome frequencies \cite{guo2017calibration}. In other words, it reduces the frequency of high-confidence but incorrect and low-confidence but correct outcomes in these models. While a calibrated model can be more accurate than an uncalibrated model, the main goal is to reduce these extreme misclassifications and align a model's confidence with the actual frequency of correctly predicted outcomes. Hence, a low-confidence judgment from a calibrated model can be deferred to another model or human reviewers and a high-confidence judgment can be accepted with lower risk of it being a false-positive, compared to an uncalibrated model. Yet it remains unclear how to reliably apply such techniques to \emph{multimodal} tasks like text-to-image (TTI) generation, which require a judge to produce outputs based on both visual and textual context. These subtleties remain relatively underexplored in current research. Moreover, prior work often assumes ideal conditions (e.g., all prompts are equally performant, prompting is \myquote{free} to perform ad infinitum), assumptions that rarely hold in real-world multimodal evaluations. Consequently, many challenges in MLLM-based evaluation remain unsolved.

In this paper, we study \emph{text-to-image} generation as an exemplar of these open-ended vision-language tasks and address the problem of producing a well-calibrated judge, ensuring that the MLLM-based judge accurately appraises its uncertainty and delivers stable, contextually coherent TTI evaluations. Achieving robust calibration in the multimodal domain not only leads to fair and trustworthy metrics for TTI tasks, but also lays the groundwork for future applications—such as selective classification~\cite{hendrickx2024machine} or partial human oversight~\cite{madras2018predict} of automated judgments—if so desired.

\xhdr{Contributions.} Put briefly, in this work we ---
\begin{compactitem}[\hspace{-2pt}•] 
\item Identify limitations of standard prompt ensemble methods when applied to multimodal evaluation, demonstrating their failure to generalize effectively to TTI judgment.
\item Propose \mmblong, a novel method incorporating image clusters to condition prompt selection, improving calibration and judgment consistency in MLLM-based TTI evaluation. 
\item Conduct extensive experiments on HPSv2 and MJBench, showing that \mmb significantly improves calibration and alignment with human preferences compared to SOTA.
\item Analyze the practical implications of \mmb, demonstrating its benefits for cost-aware evaluation pipelines, where low-confidence judgments can be selectively accepted or deferred to human reviewers for more precise review.
\end{compactitem}

\section{Related Work}
\label{sec:related}

\xhdr{(M)LLM-As-a-Judge.} (M)LLM-as-a-judge has seen much interest in recent years, serving as an economical and scalable alternative to human evaluation. Proprietary models such as GPT-4(V) have been used as general-purpose judges for various use text-only and multimodal judgments. To this end, several benchmarks have been recently proposed in both text-only evaluations, such as
LLaVA-Bench~\citep{liu2023llava}, GAVIE~\citep{liu2024mitigating}, LAMM~\citep{yin2023lamm}, and VisIT-Bench~\citep{bitton2023visitbench}, and multimodal evaluations, such as X-IQE~\citep{chen2023x}, MLLM-Bench~\citep{ge2023mllm}, ViGOR~\citep{yan2024vigor} etc. Open-source alternatives to these proprietary models have also been recently introduced \cite{xiong2024llava}, aiming to improve MLLMs in their capacity to act as judges. Our work is complementary to these developments. We seek to improve the calibration of these (M)LLMs when they're used as judges so that we can properly quantify when they are less confident in their evaluation. This helps improve the reliability, fairness, and accuracy of these models when used as judges and can also act as a filter, whereby only the less confident judgments need to be verified by humans.

\xhdr{Model Calibration.} 
We tackle model calibration, a well-established subfield in ML literature, e.g., \cite{guo2017calibration}. A calibrated model allows selective prediction of outputs  \cite{geifman2017selective, dancette2023improving}, or selectively defer \cite{madras2018predict} or abstain \cite{whitehead2022reliable} from producing an output when the model is not confident that it can produce a correct output. Contrary to our work, these works introduce various statistical and post-training interventions to improve the calibration of ML models. Our work tackles improving the calibration of black-box models, including proprietary models where we do not have access to the models' weights. 

\xhdr{Prompt Ensembles.} 
As LLMs are highly sensitive to prompt engineering \cite{jiang2020can, white2023prompt}, researchers have explored various prompt ensembling techniques to mitigate this sensitivity and improve calibration. For example, \cite{wightman2023strength, jiang2023calibrating} investigated methods to generate diverse prompts aimed at enhancing calibration, treating all prompts within the ensemble as equally important. In contrast, \citet{hou2023promptboosting, tonolini2024bayesian} assign different weights to prompts within the ensemble and optimize these weights using a validation set. Unlike \citet{hou2023promptboosting} approach, \citet{tonolini2024bayesian}  does not require modifying the prompts themselves, making it more practical for real-world applications.  To achieve this, the authors in \cite{tonolini2024bayesian} proposed Bayesian Prompt Ensembles (\bpe), a Bayesian approach to learning the varying importance of different prompts. This technique is particularly relevant when an LLM acts as a judge with multiple task instruction prompts that are assumed to be equally relevant. However, in multimodal evaluations, such assumptions might not always hold, as the relevance of a prompt can also depend on the image itself. This paper aims to address this issue and propose a more generalized solution.

\newcommand{\Dval}{\mathcal{D}_{val}}
\newcommand{\Dsup}{\mathcal{D}_{sup}}
\newcommand{\Z}{\mathcal{Z}}
\newcommand{\G}{\mathcal{G}}
\renewcommand{\P}{\mathcal{P}}
\newcommand{\R}{\mathbb{R}}

\section{Preliminaries}

\begin{figure*}[t!]
    \centering 
    \includegraphics[width=\textwidth]{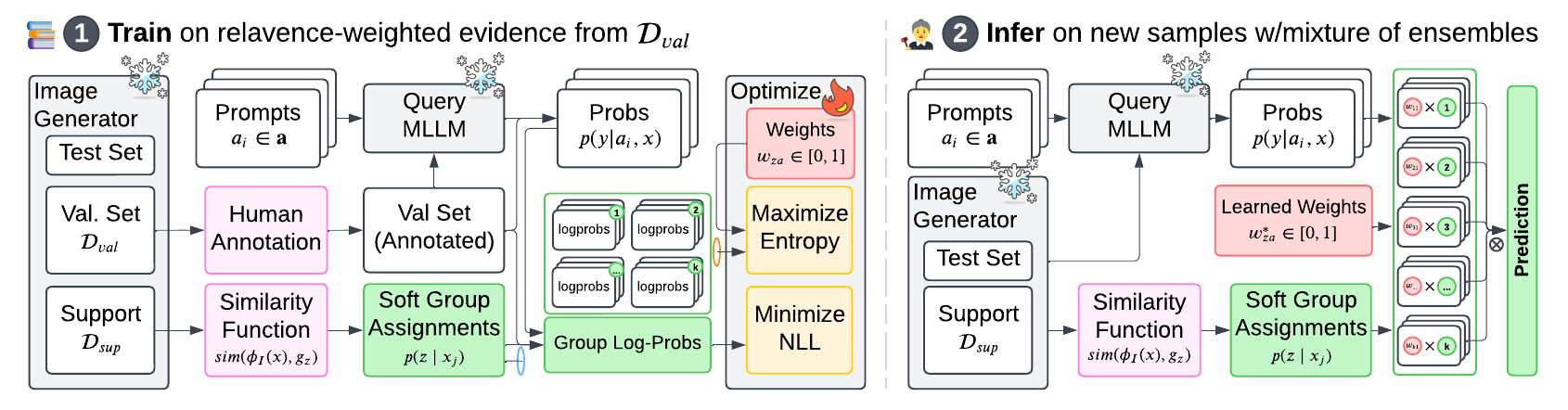}
    \caption{Overview of our \mmb for multimodal prompt ensembles. \mycirc{1} \textbf{Train}: We query an MLLM with multiple prompts and discount the resulting log-probabilities inverse to a relevance function that scores how well each image fits a learned embedding-based group. We optimize the image-conditional prompt weights by minimizing negative log-likelihood on validation data, balanced by an entropy regularizer. \mycirc{2} \textbf{Infer}: Learned prompt weights then mix MLLM outputs at test time for more accurate, calibrated predictions.}
    \label{fig:long-system}
    \vspace{0pt}
\end{figure*}

One fundamental requirement for reliable model \emph{judge} deployment is \emph{calibration}: the idea that a model's predicted probabilities should accurately reflect the true likelihood that its predictions are correct. Given an input $x$ (\eg, an image pair), a predicted label $\hat{y}$, and a ground truth label $y^*$ (\eg, a known preference), a well-calibrated model satisfies:
\begin{align}
P(y^* = \hat{y} \mid f(\hat{y} \mid x) = p) = p
\end{align}
where $f(\hat{y}\mid x)$ denotes the model's predicted probability for label $\hat{y}$ given input $x$. In other words, if a model outputs a 90\% probability of being correct on some sample, we should find that---over many samples assigned that same 90\% confidence---it is indeed correct about 90\% of the time.

\subsection{Bayesian Prompt Ensembles (\bpe)}

Our work builds on \emph{Bayesian Prompt Ensembles} (\bpe) \cite{tonolini2024bayesian}, originally proposed to improve calibration in black-box language models. Although \bpe was introduced for purely textual (NLP) classification tasks, our goal is to generalize its principles to \emph{multimodal} model evaluation---where images, prompts, and model outputs all interact. To apply this framework to our MLLM setting, we assume:
\begin{compactitem}[\hspace{-2pt}•] 
    \item A set of $N$ semantically equivalent \emph{task prompts} $\boldsymbol{a} = \{a_1,\dots,a_N\}$. Each prompt describes the same classification or preference task (e.g., \myquote{Which of these two images is more realistic?}) in slightly different wording.
    \item A small validation set $\Dval{=}\{(x_j, y^*_j)\}_{j=1}^M$. Here, $x_j$ are inputs (which could be text, images, or both), and $y_j^*$ are ground truth labels (e.g., human preferences).
    \item A fixed black-box model (e.g., an MLLM) providing class probabilities $p(y|x,a)$ given input $x$ and a prompt $a$.
\end{compactitem}

\xhdr{Prompts as Latent Variables.} \bpe treats each prompt $a$ as a latent variable in a Bayesian sense. For an (M)LLM-based classifier, the desired predictive distribution is:
\begin{align}
p(y \mid x, \Dval) \;=\;\int p\bigl(y \mid x,a\bigr)\,p\bigl(a \mid \Dval\bigr)\,da.
\end{align}
Since $p(y|x,a)$ is fixed once the (M)LLM and prompt are chosen, the goal is to approximate the posterior $p(a|\Dval)$.

\xhdr{Variational Inference.} \bpe introduces a variational distribution $q(a)$ to approximate $p(a|\Dval)$, minimizing the KL:
\begin{align}
q^*(a) \;=\; \operatorname*{arg\,min}_{q(a)} KL\bigl[q(a)\,\|\,p(a\mid \Dval)\bigr].
\end{align}
By standard variational arguments (see, e.g., \cite{graves2011practical}),
\begin{align}
q^*(a)
\;=\;
\operatorname*{arg\,max}_{q(a)}\;\Bigl(
\mathbb{E}_{q(a)}[\log &p(\boldsymbol{y}^*\mid \boldsymbol{x},a)] \label{eq:bpe_elbo}
\\ & \;-\;
KL[\,q(a)\,\|\,p(a)\,]
\Bigr). \notag 
\end{align}
Intuitively, $q(a)$ places higher density on prompts that explain the validation data well.

\xhdr{Discrete Reparameterization.} In practice, we only have a finite set $\boldsymbol{a} = \{a_i\}_{i=1}^N$. \bpe thus represents $q(a)$ with discrete weights $w_a{\geq}0$ such that $\sum_a w_a{=}1$ and $w_a{=}q(a)/NC$. Assuming a uniform constant prior $p(a)\approx C$, this yields:
\begin{align} \label{eq:bpe-importance}
\operatorname*{arg\,max}_{\bf w}
\sum_{a} w_a \Bigl[\sum_{j=1}^M& \log p\bigl(y_j^*\mid x_j, a\bigr)\Bigr]
\\ & \;-\;
\sum_{a} w_a \,\log w_a, \notag
\end{align}
where the first term rewards prompts whose likelihood on $\Dval$ is high, and the second term is an entropy term that prevents the solution from collapsing onto a single prompt unless it decisively outperforms the rest.

\xhdr{Inference.} Once the weights $w_a^*$ are learned, the final class probability for a new sample $x$ becomes:
\begin{align}
p(y\mid x)
\; \approx\;
\sum_{a}
w_a^*
\;p\bigl(y\mid x, a\bigr).
\end{align}
Hence \bpe combines the model outputs from multiple task prompts, reweighting them for better calibration.

\subsection{Limitations \& Our Multimodal Generalization}
\bpe focuses on textual classification, assuming all task prompts are \emph{equally relevant a priori}. Multimodal judging tasks, however, may demand different prompts for different \emph{types of images}: for example, a prompt that references lighting or artistic style might be more reliable for photos than for abstract digital art. Consequently, \bpe can be suboptimal when the best prompt for a given validation sample varies by \emph{image category} or other visual attributes.

In the remainder of this paper, we propose a \emph{multimodal generalization} of \bpe that conditions on an image embedding to cluster or group related images. Each group can then favor the prompts best suited for that group. Our method thus preserves \bpe's variational formulation but learns more \emph{image-specific} weights, significantly improving both accuracy and calibration in MLLM-based evaluation.

\section{Multimodal Bayes. Prompt Ensembles}
\label{sec:mmb}

We now propose \mmblong (\mmb), a technique for learning \emph{image-aware} prompt weights that generalize \bpe into the multimodal domain. Our key idea is model an underlying group structure based on \emph{image embeddings} --- allowing the model to learn group-specific prompt weights and individual samples to be classified by their combination based on group affinity. This helps address scenarios where different prompts may be more reliable for certain types of images. See \cref{fig:long-system} for a high-level graphical overview.

\xhdr{Soft Image Grouping.}
To model this group structure, we presume the image space can be partitioned into $K$ groups which may each have different prompt weights that are more appropriate for the image in group. More formally,  we introduce random variables $z{\mid}x$ to denote the membership of image $x$ in group $z \in \{1,...,K\}$. To realize this grouping, we assume access to an unlabeled image set $\Dsup$ drawn i.i.d from the image generator we seek to evaluate (\ie $\G: * \to x$) and an image embedding function $\phi_I: X \to \R^d$ (for instance, via a pretrained image encoder). Applying a grouping algorithm (e.g., k-means) to image embeddings, we can partition $\Dsup$ into $K$ groups $g_1,...,g_K$. For a similarity function $\mbox{sim}(\phi_I(x), g_i)$ between the image embedding of $x$ and group $g_z$, we can define the probability of $z\mid x$ as $p(z|x) \propto exp({\mbox{sim}\left(\phi_I(x), g_z\right)/\tau})$,
where $\tau$ is a temperature scale hyperparameter. For $\tau\to0$, $p(z|x)$ approaches a one-hot hard assignment and a uniform distribution for $\tau\to\inf$. This distribution can be viewed as a soft assignment of $x$ to each of the $K$ groups. 

\subsection{Learning Objective}
Given this group structure, we can derive the evidence lower bound for the log likelihood of $\Dval$. To start, the log likelihood for a given point $x,y$ can be written as:
\begin{eqnarray}
\log p(y|x) 
 &{=}& \log \sum_{a} \sum_{z} p(y{\mid}x,a) p(a{\mid}z) p(z{\mid}x)   
\end{eqnarray}
where conditional independencies $a{\perp}x{\mid}z$ and $y{\perp}z{\mid}a$ are applied. For these, recall that we assume prompt weights are determined entirely by group assignment and our underlying model's output depends only on the input and prompt. Introducing a variational distribution $q(a|z)$ to model the unknown group-specific prompt weights and following standard manipulations yields an evidence lower bound of $\log p(y|x)$ equal to
\begin{eqnarray}
\log p(y|x)&{\geq}&\mathbb{E}_{p(z|x)} \left[\vphantom{\sum}~~\mathbb{E}_{q(a|z)}\log p(y|x,a) \right.\nonumber\\&&~~~~~~~~~~~~~~\left.\vphantom{\sum}-\mbox{KL}(q(a|z) || p(a|z))~~\right]~~~           
\end{eqnarray}

Note that the terms inside $\mathbb{E}_{p(z|x)}$ mirror \bpe in \cref{eq:bpe_elbo} but now have group-specific weights $q(a{\mid}z)$ and priors $p(a{\mid}z)$. As we've defined it, the expectation over $p(z{\mid}x)$ serves to weigh point $x$'s per-group contribution based on similarity to each group. See \cref{sec:supp-derivation} for full derivation. \looseness=-1

Setting a uniform prior $p(a{\mid}z)~\forall z$ and parameterizing the variational distribution $q(a|z)$ via learnable weights $w_{za}$, the training objective for our \mmb formulation is then to find weights which maximize the following:
\newcommand{\shim}{\vphantom{\left[ \sum_{a}\right.}}
\begin{eqnarray}
    \sum_{j=1}^M \sum_z \overbrace{p(z|x_j)\shim}^{\mbox{\footnotesize\shortstack{Soft Group\\Assignment}}}\overbrace{\shim \left[\sum_a w_{za} \log p(y_j^*|x_j,a)\right.}^{\mbox{\footnotesize\shortstack{Per-Group \\Log Likelihood}}}\nonumber\\ 
     \underbrace{\left. - \sum_a w_{za} \log w_{za}~~ \right]}_{\mbox{\footnotesize \shortstack{Per-Group Entropy\\Regularizer}}}
     \label{eq:mmb-obj}
\end{eqnarray}

With our complete objective written, it is worth revisiting our temperature hyperparameter $\tau$ used in defining $p(z{\mid}x)$. We note that extreme settings will map the above objective to either (i) independent \bpe per group when all $p(z{\mid}x_i)$ become one-hot ($\tau\to0$) or (ii) one single over-parameterized \bpe for all data when all $p(z{\mid}x_i)$ become uniform ($\tau\to\inf$). For intermediate settings, data points can selective share partial membership across image groups based on their semantic or visual similarity.

\subsection{Inference}
Solving Eq.~\ref{eq:mmb-obj} for $w_{za}^*$, enables us to evaluate new inputs $x$ as an expectation over group assignments and prompts,
\begin{equation}
\label{eq:mmb-infer}
    p(y\mid x)\approx
    \sum_{z}p(z|x)\sum_{a}w_{za}^*\;p\bigl(y{\mid}x,a_i\bigr),
\end{equation}
\noindent where $p(z|x)$ is computed based on $x$'s group similarities. 

\begin{table*}[!ht]
\centering
\begin{tabular}{l@{\hspace{-0.5em}}l@{}cccc>{\columncolor[HTML]{F0F0F0}}ccccc>{\columncolor[HTML]{F0F0F0}}ccccc>{\columncolor[HTML]{F0F0F0}}c}
\toprule
\multirow{2}{*}[-0.8em]{\rotatebox{50}{\scriptsize prompts}} & 
\multirow{2}{*}[-0.8em]{\rotatebox{50}{\scriptsize samples}} & 
\multicolumn{5}{c}{Expected Calibration Error ($\downarrow$)} & 
\multicolumn{5}{c}{Max Calibration Error ($\downarrow$)} &
\multicolumn{5}{c}{AUC Precision-Recall ($\uparrow$)} \\
\cmidrule(l{3pt}r{3pt}){3-7}
\cmidrule(l{3pt}r{3pt}){8-12}
\cmidrule(l{3pt}r{3pt}){13-17}
& 
& 
\makecell{\textbf{\footnotesize \std} \\[-5pt] {\tiny (single)}} & 
\makecell{\textbf{\footnotesize \bst} \\[-5pt] {\tiny (single)}} & 
\makecell{\textbf{\footnotesize \avg} \\[-5pt] {\tiny (ensemble)}} & 
\makecell{\textbf{\footnotesize \bpe} \\[-5pt] {\tiny (ensemble)}} & 
\makecell{\textbf{\footnotesize \mmb} \\[-5pt] {\tiny (ensemble)}} &
\makecell{\textbf{\footnotesize \std} \\[-5pt] {\tiny (single)}} & 
\makecell{\textbf{\footnotesize \bst} \\[-5pt] {\tiny (single)}} & 
\makecell{\textbf{\footnotesize \avg} \\[-5pt] {\tiny (ensemble)}} & 
\makecell{\textbf{\footnotesize \bpe} \\[-5pt] {\tiny (ensemble)}} & 
\makecell{\textbf{\footnotesize \mmb} \\[-5pt] {\tiny (ensemble)}} &
\makecell{\textbf{\footnotesize \std} \\[-5pt] {\tiny (single)}} & 
\makecell{\textbf{\footnotesize \bst} \\[-5pt] {\tiny (single)}} & 
\makecell{\textbf{\footnotesize \avg} \\[-5pt] {\tiny (ensemble)}} & 
\makecell{\textbf{\footnotesize \bpe} \\[-5pt] {\tiny (ensemble)}} & 
\makecell{\textbf{\footnotesize \mmb} \\[-5pt] {\tiny (ensemble)}} \\
\midrule
\multirow{4}{*}{5}
    & 5   & .238 & .155 & .155 & .127 & \textbf{.113} & .399 & .351 & .286 & .305 & \textbf{.245} & .731 & .812 & \textbf{.835} & \textbf{.830} & \textbf{.835} \\
    & 10  & .239 & .132 & .155 & .121 & \textbf{.108} & .406 & .320 & .286 & .304 & \textbf{.239} & .731 & \textbf{.841} & .835 & \textbf{.847} & \textbf{.838} \\
    & 20  & .243 & .130 & .155 & .120 & \textbf{.108} & .411 & .313 & .286 & .308 & \textbf{.241} & .724 & \textbf{.842} & .835 & \textbf{.849} & \textbf{.838} \\
    & 50  & .261 & .122 & .155 & .121 & \textbf{.107} & .424 & .310 & .286 & .307 & \textbf{.236} & .708 & \textbf{.853} & .835 & \textbf{.853} & \textbf{.839} \\
 \arrayrulecolor{black!20}\midrule
\multirow{4}{*}{10}
    & 5   & .271 & .150 & .142 & .121 & \textbf{.092} & .409 & .346 & .250 & .291 & \textbf{.201} & .694 & .818 & \textbf{.844} & .835 & \textbf{.844} \\
    & 10  & .260 & .134 & .142 & .120 & \textbf{.095} & .401 & .328 & .250 & .293 & \textbf{.196} & .702 & \textbf{.842} & \textbf{.844} & .837 & \textbf{.844} \\
    & 20  & .260 & .127 & .142 & .114 & \textbf{.091} & .401 & .317 & .250 & .285 & \textbf{.189} & .702 & \textbf{.847} & \textbf{.844} & \textbf{.848} & \textbf{.845} \\
    & 50  & .267 & .121 & .142 & .116 & \textbf{.088} & .408 & .301 & .250 & .289 & \textbf{.188} & .696 & \textbf{.854} & \textbf{.844} & .851 & \textbf{.845} \\
\midrule
\multirow{4}{*}{20}
    & 5   & .263 & .153 & .133 & .111 & \textbf{.080} & .422 & .342 & .210 & .274 & \textbf{.172} & .716 & .812 & \textbf{.849} & .841 & \textbf{.847} \\
    & 10  & .265 & .135 & .133 & .117 & \textbf{.082} & .422 & .324 & .210 & .288 & \textbf{.169} & .713 & .835 & \textbf{.849} & .841 & \textbf{.848} \\
    & 20  & .270 & .126 & .133 & .114 & \textbf{.080} & .426 & .311 & .210 & .279 & \textbf{.160} & .708 & \textbf{.847} & \textbf{.849} & .844 & \textbf{.848} \\
    & 50  & .267 & .117 & .133 & .113 & \textbf{.076} & .422 & .291 & .210 & .275 & \textbf{.154} & .708 & \textbf{.855} & \textbf{.849} & .851 & \textbf{.849} \\
\arrayrulecolor{black}\bottomrule
\end{tabular}
\caption{Expected Calibration Error (ECE) and Max Calibration Error (MCE) are shown (lower is better $\downarrow$), along with AUC Precision-Recall (higher is better $\uparrow$) on HPSv2~\cite{wu2023human}. We compare 
\textbf{\std (single)}---a random single prompt, 
\textbf{\bst (single)}---the single prompt with highest validation accuracy, 
\textbf{\avg (ensemble)}---an unweighted average, 
\textbf{\bpe (ensemble)}---the current state of the art, 
and \textbf{MMB (ensemble)}---our proposed method. 
\textbf{Bold} entries are either the \bst score or not significantly different from the \bst at ${\geq}95\%$ confidence via a permutation test. 
We account for Type~I error inflation across multiple tests per metric using the Benjamini--Yekutieli FDR procedure~\citep{benjamini2001control}.}
\label{tab:big-res-model}
\vspace{-0.7em}
\end{table*}

\section{Experimental Setup}
\label{sec:experiments}
We conduct experiments on two contemporary benchmarks:

\xhdr{HPSv2: Discriminative Power and Calibration \cite{wu2023human}.} HPSv2 is a large-scale dataset capturing human preferences among images generated from the same textual prompts; it encompasses $\sim\!\!800{k}$ preference choices over $\sim\!\!430{k}$ images. Of these, $400$ groups (each containing 9 images) serve as a test set, and $108k$ groups (each containing 4 images) comprise the training pool. We focus on pairwise preferences drawn from these groups. 
For \textbf{calibration} (\ie, learning our ensemble weights), we randomly select a small number of pairwise comparisons—one per training group as needed—ensuring that each validation sample is distinct. We denote this validation set by $\Dval$ and from the remaining training data for $\Dsup$. After calibration, we \textbf{evaluate} the final models on all $\smash{\binom{9}{2}}\times400 = 14.4k$ pairwise comparisons from the test set.
We systematically explore several experimental factors, including the number of prompts used ($\boldsymbol{a};$ 5, 10, or 15) and the number of validation samples ($\Dval$; 5, 10, 20, or 50). Our support set ($\Dsup$) is always composed of $256{\times}K$ samples, where $K$ denotes the number of groups used in \mmb. Across each configuration, we repeat experiments with multiple random seeds for training (3), data sampling (50), and clustering (5) for a total of $52.2k$ unique experimental configurations. This yields a broad factorial design allowing for thorough comparisons of calibration and discriminative ability. We provide a summary of the experimental configurations in \cref{sec:suppliment}.

\xhdr{MJBench: Visually Salient Human Social Bias \cite{chen2024mj}.}
MJBench-Bias is a targeted evaluation set for measuring demographic biases in multimodal judge models. It comprises images of subjects from diverse backgrounds (\eg, different ages, genders, or socioeconomic statuses) with prompts describing occupations or educational pursuits. The goal is to assess whether a judge model’s scoring or ranking of how well an image aligns with a prompt is free from systematic demographic bias. Because MJBench-Bias provides pools of similar images per prompt—rather than pairs—and lacks a standard training set, we construct one to facilitate experimentation. Specifically, we create a lower-preference variant of each image by applying aesthetically degrading transformations (such as extreme contrast, motion blur, brightness shifts, random occlusions, and noise). We then form an artificial training set by pairing each original image with its transformed counterpart, chosen from images generated under the same prompt. We adopt a 10-fold, leave-one-group-out procedure stratified by prompt: in each fold, the remaining nine folds serve as the validation ($\Dval$) and support ($\Dsup$) sets—containing distorted pairs—while the held-out fold is reserved for testing. On the test set, we generate all image pairs sharing a caption; under the assumption of no bias, neither image should be more or less preferable. Consequently, we argue that an unbiased model in this scenario should predict with the lowest possible confidence—50\%. See \cref{sec:supp-synthetic} for synthetic preferences examples.

\xhdr{Models.}
Throughout all experiments, we select GPT-4o \cite{hurst2024gpt} as our \emph{judge} model. GPT-4o is a state-of-the-art, closed-source MLLM that is frequently employed in multimodal judge scenarios due to its high performance. We consider text-conditioned image generators $\mathcal{G}\!: \text{text} \times \text{noise}\!\to\!\text{image}$ as the underlying image producers for our datasets dataset which can be used to generate $\Dsup$. The embedding function $\phi_I$ is implemented using a pretrained CLIP-ViT-B16 \cite{radford2021clip}, chosen for its strong alignment with natural language and visual content understanding. To form our K-group relevance function $\mathcal{Z}$, we perform spherical k-means clustering on image embeddings via FAISS \citep{johnson2019faiss} with cosine similarity based distance between image pairs $(x_1,x_2)$:
\[
dist(x_1,x_2) = 1 - \phi(x_1)^T\phi(x_2)\ /\ \lVert \phi(x_1) \rVert \lVert \phi(x_2) \rVert
\]
We take the cosine similarity between each image and the $K$ cluster centroids as our similarity function $\mbox{sim}(\cdot, \cdot)$.

\begin{table}[t!]
\footnotesize
\centering
\begin{tabular}{llcccccc}
\toprule
& 
\textbf{Model}             &
\textbf{NLL$_\downarrow$}  & 
\textbf{Brier$_\downarrow$}&
\textbf{Kappa$_\uparrow$}  &
\textbf{Acc$_\uparrow$}    &
\textbf{ROC$_\uparrow$}    &
\textbf{F1$_\uparrow$}      \\
\midrule
\multirow{2}{*}{\rcnt{\scriptsize Single}}
& {\std} &        1.002  &         .271  &         .315  &         .667  &         .780  &          .526  \\
& {\bst} &         .618  &         .152  &         .610  &         .812  &    \textbf{.897}  &          \textbf{.763}  \\
 \arrayrulecolor{black!20}\midrule
\multirow{3}{*}{\rcnt{\scriptsize Ensemble}}
& {\avg} & .473 &          .151  &         .600  &         .804  &  \textbf{.897}  & \textbf{.764} \\
& {\bpe} &          .547  &          .147  &         .602  &         .810  & \textbf{.897}  &         .753  \\
& {\mmb} &  \textbf{.430}  & \textbf{.135} & \textbf{.627} & \textbf{.820} & \textbf{.900} & \textbf{.778} \\
 \arrayrulecolor{black}\bottomrule
\end{tabular}
\caption{Performance on HPSv2 \cite{wu2023human}
showing NLL and Brier score (lower is better $\downarrow$) alongside Kappa, Accuracy (Acc), ROC-AUC (ROC), and F1 (higher is better $\uparrow$). We compare two single-prompt baselines (\std and \bst) against three ensemble methods: \avg (unweighted average), \bpe (SOTA), and \mmb (ours). Methods insignificantly different from the best method by column at 95\% confidence from a permutation test are in \textbf{bold}. We control for Type~I error from multiple testing with the Benjamini--Yekutieli FDR procedure~\cite{benjamini2001control}. 10 prompts, 20 validation samples.
}
\label{tab:small-res-model}
\end{table}

\xhdr{Baselines.}
We benchmark our proposed \mmb approach against two single prompt and two ensemble baselines:
\begin{compactitem}[\hspace{-2pt}•] 
\item \emph{Standard}: A single randomly chosen prompt.
\item \emph{Best}: The single prompt with greatest $\Dval$ accuracy.
\item \emph{Average}: A simple average over all available prompts.
\item \emph{BPE}: A state-of-the-art Bayesian method originally developed for text-only prompt ensembling \citep{tonolini2024bayesian}.
\end{compactitem}
We draw all prompts from a pool of 100 diverse and semantically equivalent instructions, combining manual definitions, structured templates, and LLM-driven rephrasings. See \cref{sec:supp-prompts} for further prompt generation details.

\xhdr{Metrics.} Following standard practice, we use the \emph{Expected Calibration Error (ECE)} and \emph{Maximum Calibration Error (MCE)} \cite{guo2017calibration} to measure calibration. In short, these methods summarize the discrepancy between model confidence and actual correctness across bins in a reliability diagram. Discriminative ability and alignment with human annotations are measured using \emph{Cohen’s Kappa (Kappa)} \cite{cohen1960kappa}, \emph{ROC-AUC} \cite{bradley1997auc}, and \emph{AUC Precision-Recall (PR)} \cite{davis2006relationship}, along with traditional metrics such as \emph{Accuracy (Acc)}, \emph{F1-score}, Brier score~\cite{brier1950score} and test-set NLL~\cite{hastie2001nll}. To succinctly summarize our methods’ discriminative power and calibration across conditions, we present detailed results primarily for ECE, MCE, and AUC-PR. Results for additional metrics are fully reported for one representative setting (10 prompts, 20 validation samples). For MJBench, we report average confidence on our synthetic equal-preference task by method.

\section{Results}
\label{sec:results}

\begin{figure*}[t]
    \centering
    \includegraphics[width=\textwidth]{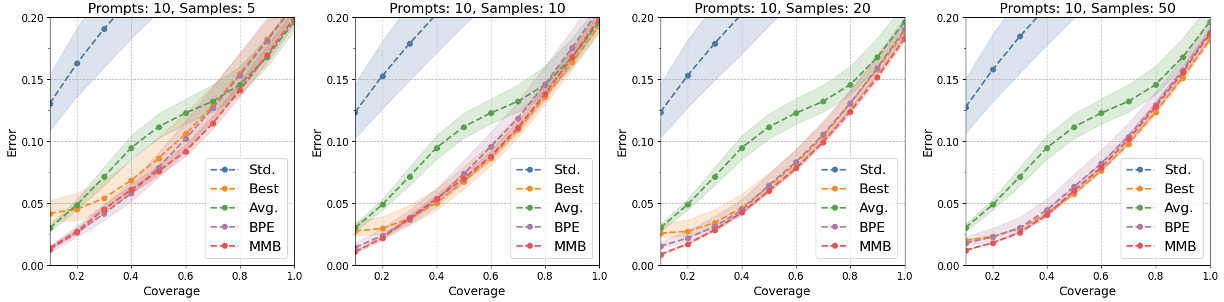}
    \caption{Error--coverage curves on HPSv2~\citep{wu2023human} with 10 prompts across varying numbers of validation samples (5, 10, 20, 50). Each curve represents an average over multiple runs, with 95\% confidence intervals (shaded regions) from bootstrapped sampling of the mean. Our \mmb approach consistently achieves the lowest error across coverage levels and displays the narrowest intervals, indicating more stable and reliable performance. We show similar curves across additional experimental configurations in \cref{sec:supp-additional-results} \cref{fig:big-coverage}.}
    \label{fig:small-coverage}
    \vspace{-11pt}
\end{figure*}

For each $nprompt \times nsample$ prompt-ensemble configuration, we perform a paired permutation test on the mean difference in performance between the best-performing model and each other method. To control for Type-I error inflation from multiple testing, we apply the Benjamini--Yekutieli False Discovery Rate correction~\cite{benjamini2001control} on a per-metric basis. As shown in \cref{tab:big-res-model}, \mmb outperforms both single-prompt approaches (\std and \bst) and existing prompt ensembling techniques (\avg and \bpe) in our multimodal setup overall. Looking deeper into the specific 10 prompt 20 sample configuration, \cref{tab:small-res-model} shows additional evidence that \mmb consistently delivers improved calibration (\eg, ECE, Brier) and stronger discriminative metrics (\eg, AUC-PR), across a variety of measures. Notably, \mmb performs well even when the number of prompts and validation samples is low, though performance generally improves as we increase either of those information sources.

\begin{table}[t!]
\centering
\footnotesize
\begin{tabular}{rccccccc}
\toprule
\textbf{K}             &
\textbf{ECE$_\downarrow$} &
\textbf{NLL$_\downarrow$}  & 
\textbf{Brier$_\downarrow$}&
\textbf{Kappa$_\uparrow$}  &
\textbf{Acc$_\uparrow$}    &
\textbf{ROC$_\uparrow$}    &
\textbf{F1$_\uparrow$}      \\
\midrule
  4 & \textbf{.090} &         .432  &          .140  &         .627  &         .819 &         .899 &  .777  \\
  8 & \textbf{.090} & \textbf{.430}  & \textbf{.135} &         .627  &         .820 &         .899 &  .778 \\
 16 & .091 & \textbf{.430}  & \textbf{.135} & \textbf{.627} & \textbf{.820} &        .900 &  \textbf{.779} \\
 32 & .091 & \textbf{.430}  & \textbf{.135} &         .627  &         .820 &         .900 &  .778 \\
 64 & .091 & \textbf{.430}  & \textbf{.135} &         .627  &         .820 & \textbf{.900} & .778  \\
\bottomrule
\end{tabular}
\caption{Effect of varying \mmb cluster count $K$ on HPSv2~\citep{wu2023human} using 10 prompts and 20 validation samples, showing NLL and Brier score (lower is better~$\downarrow$) alongside Kappa, Accuracy (Acc), ROC-AUC (ROC), and F1 (higher is better~$\uparrow$). Performance improves slightly as $K$ increases, then saturates at around 32 clusters. $K$'s insignificantly different from the best by column at 95\% confidence from a permutation test are in \textbf{bold}. Type~I error from multiple testing controlled with the Benjamini--Yekutieli~\cite{benjamini2001control} method.
}
\label{tab:small-res-cluster}
\end{table}

\begin{figure*}[t]
    \centering 
    \captionsetup[subfigure]{format=hang, width=0.98\linewidth} 
    \begin{subfigure}[b]{0.33\textwidth}
        \centering\includegraphics[width=\textwidth, trim=0 55 0 0, clip]{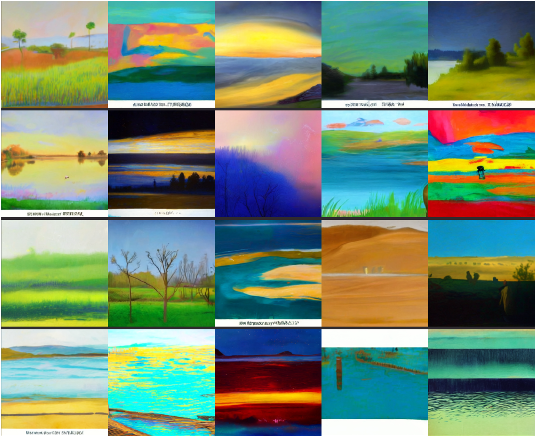}
        \caption{\myquote{You are a \textbf{landscape artist} skilled in assessing lighting, color and composition [...]}}
        \label{fig:small-qualitative-landscape}
    \end{subfigure}
    \begin{subfigure}[b]{0.33\textwidth}
        \centering
    \captionsetup[subfigure]{format=hang, width=\linewidth} 
        \includegraphics[width=\textwidth, trim=0 55 0 0, clip]{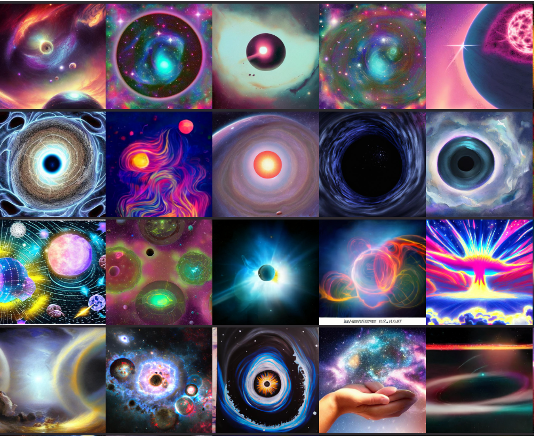}
        \caption{\myquote{You are a \textbf{graphic designer} with experience in visual clarity and technical image quality [...]}}
        \label{fig:small-qualitative-graphic}
    \end{subfigure}
    \begin{subfigure}[b]{0.33\textwidth}
        \centering
    \captionsetup[subfigure]{format=hang, width=\linewidth} 
        \includegraphics[width=\textwidth, trim=0 55 0 0, clip]{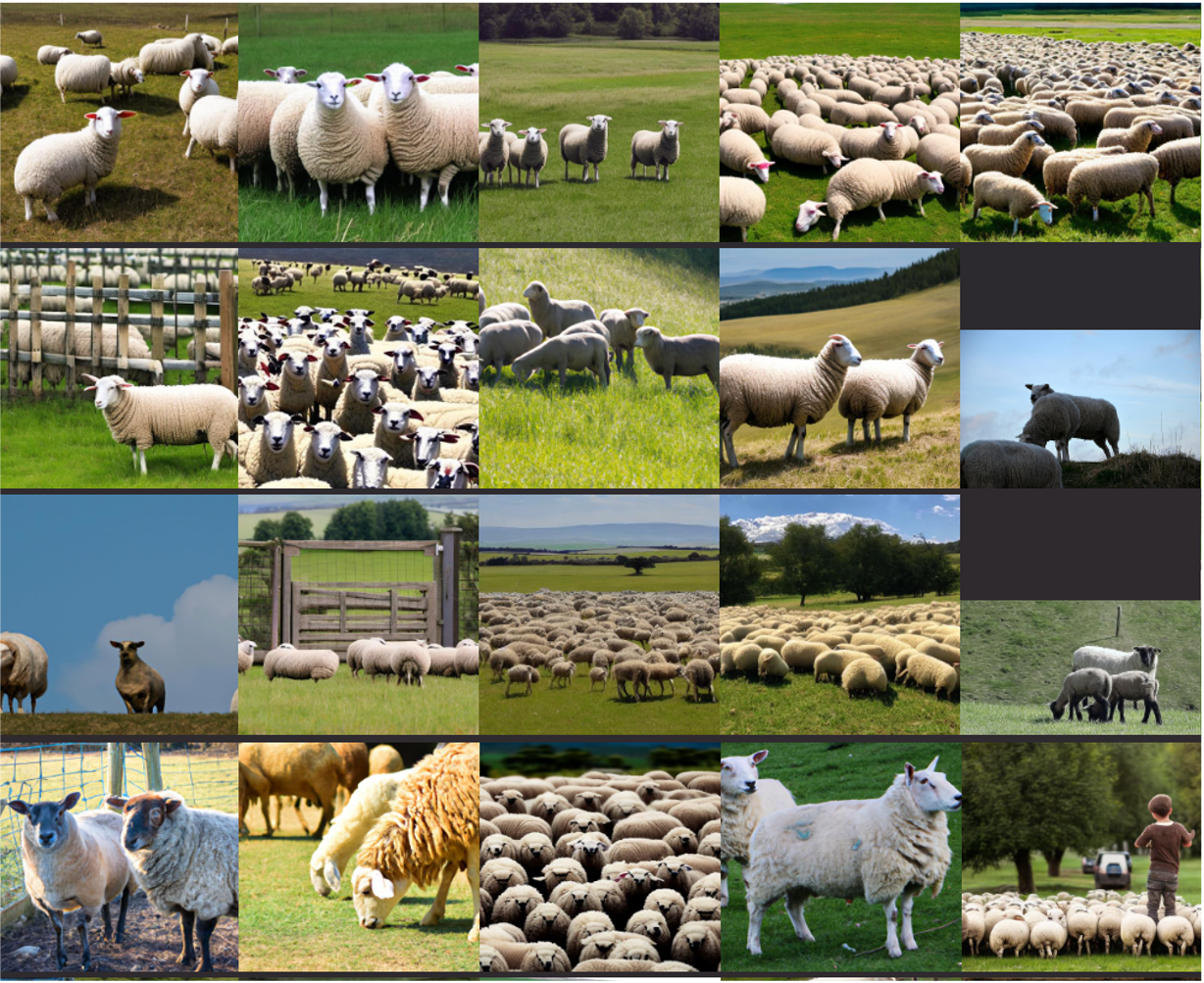}
        \caption{\myquote{You are a \textbf{photographer} skilled in assessing lighting, focus, and exposure [...]}}
        \label{fig:small-qualitative-sheap}
    \end{subfigure}
    \caption{Clusters from \mmb's grouping, each matched to its best-performing prompt by largest weight in $\boldsymbol{w_{k}}$. In (a), a prompt posing the user as a \myquote{landscape artist} is weighted highest for images featuring fields and skies, while in (b), a \myquote{graphic designer} prompt better suits the cosmic artwork. This demonstrates \mmb's prompt utilization aligning with each cluster’s style. See \cref{sec:supp-qualitative} for more. \vspace{-1em}}
    \label{fig:small-qualitative}
\end{figure*}

\xhdr{Qualitative Observations.} In order to understand better \emph{why} MMB is successfully, we visualize several clusters alongside their highest weighted prompt when running MMB in exceptionally favorable settings. That is to say, with many clusters (64) and a large number of validation samples (200). In this scenario, most clusters which are meaningful tend towards the best prompt for that cluster as entropy is essentially dropped over the increasing NLL sum. Interestingly, we find that there is good correspondence between the personas mentioned in the best performing prompts and the images they're judging. For example in \cref{fig:small-qualitative}, our pastel drawings of fields and plains are best judged by the landscape artist persona, and the vibrant sci-fi renderings of galaxy and space are best rated by the graphic designer persona. See \cref{sec:supp-qualitative} for additional examples.

\xhdr{Comparison Across Clusters.} 
\cref{tab:small-res-cluster} further examines the effect of the number of clusters in \mmb. Even a relatively small number of clusters (e.g., 8 or 16) already confers most of the performance advantage, beyond which performance starts to saturate. Hence, \mmb is robust to a range of cluster granularities and does not require an excessively large $K$ to achieve benefits. These results also suggest that excessively values of $k$ (${>}{64}$) may degrade performance.

\begin{table}[t!]
\centering
\footnotesize
\begin{tabular}{r@{\hspace{0.5em}}rccccc}
\toprule
{prompts} & 
{samples} & 
\makecell{\textbf{\footnotesize \std} \\[-5pt] {\tiny (single)}} & 
\makecell{\textbf{\footnotesize \bst} \\[-5pt] {\tiny (single)}} & 
\makecell{\textbf{\footnotesize \avg} \\[-5pt] {\tiny (ensemble)}} & 
\makecell{\textbf{\footnotesize \bpe} \\[-5pt] {\tiny (ensemble)}} & 
\makecell{\textbf{\footnotesize \mmb} \\[-5pt] {\tiny (ensemble)}} \\
\midrule
\multirow{4}{*}{5} 
    & 5   & .838 & .830 & \textbf{.683} & .738 & \textit{.726} \\
    & 10  & .838 & .831 & \textbf{.683} & .739 & \textit{.727} \\
    & 20  & .838 & .831 & \textbf{.683} & .742 & \textit{.727} \\
    & 50  & .838 & .832 & \textbf{.683} & .744 & \textit{.728} \\
 \arrayrulecolor{black!20}\midrule
\multirow{4}{*}{10}
    & 5   & .839 & .831 & \textbf{.656} & .713 & \textit{.704} \\
    & 10  & .839 & .831 & \textbf{.656} & .714 & \textit{.705} \\
    & 20  & .839 & .831 & \textbf{.656} & .715 & \textit{.705} \\
    & 50  & .839 & .832 & \textbf{.656} & .720 & \textit{.705} \\
\midrule
\multirow{4}{*}{20}
    & 5   & .834 & .831 & \textbf{.654} & .711 & \textit{.702} \\
    & 10  & .834 & .831 & \textbf{.654} & .711 & \textit{.703} \\
    & 20  & .834 & .832 & \textbf{.654} & .712 & \textit{.703} \\
    & 50  & .834 & .832 & \textbf{.654} & .715 & \textit{.703} \\
 \arrayrulecolor{black}\bottomrule
\end{tabular}
\caption{Average confidence (lower is better $\downarrow$) on our synthetic MJBench no-preference test split. The best and second best models in each row are in \textbf{bold} and \textit{italics}, respectively.
}
\label{tab:small-res-bias}
\end{table}

\xhdr{MJBench Results.}
\cref{tab:small-res-bias} reports the average confidence of each method on our synthetic MJBench-Bias no-preference test split. The Average ensemble consistently produces the lowest confidence scores, suggesting it is the most cautious judge in scenarios where no preference should exist. However, this comes at the cost of lower overall performance in discriminative tasks, as seen in our main results. In contrast, \mmb achieves the second-lowest confidence scores while maintaining strong overall performance, making it the most balanced method. This suggests that \mmb is the most fair method that remains performant enough for real-world use, mitigating overconfidence in ambiguous cases without sacrificing accuracy in general evaluation tasks.

\section{Discussion}
\label{sec:discussion}

One motivation for a well-calibrated MLLM-based judge is to enable \emph{selective} or \emph{cost-aware} evaluation pipelines. For instance, a system can elect to trust automatically produced ratings only for samples on which it is sufficiently confident, and refer low-confidence cases for human review. In \cref{fig:small-coverage}, we show coverage-error curves to illustrate how \mmb behaves in such a scenario. Here, \emph{coverage} is the fraction of examples whose confidence surpasses a chosen threshold. \emph{Error} is the misclassification rate among just those covered samples. Ideally, as coverage increases, accuracy remains high. We see that \mmb consistently yields the lowest error across coverage levels and exhibits narrower confidence intervals (shaded regions in the plot) than either single prompts or prior ensemble methods. This consistency implies that one can safely raise the confidence threshold (thus covering more samples automatically) without a large spike in error. Notably, error-coverage curves are created independent of the confidence thresholds which would generate each point. A well-calibrated model allows a developers to effectively select confidence thresholds which map to points on the error-coverage diagram due to alignment between the y-axis error and confidence values. Together, this means \mmb produces desirable error-coverage curves, and allows for trustworthy threshold selection to align with developer needs. See \cref{sec:supp-additional-results} for additional plots.

\xhdr{Behavior Under Extreme Settings.}
We note special cases reducing the behavior of \mmb to a simpler alternative:
\begin{compactitem}[\hspace{-2pt}•] 
    \item \emph{No validation data} ($|\Dval|=0$): \mmb reduces to an average ensemble, without any ground-truth labels to distinguish prompt performance, the weights remain uniform.
    \item \emph{Excessive validation data} ($|\Dval|\to \infty$): 
    \mmb converges on the best single prompt for each cluster, provided the validation set covers a wide variety of image content.
    \item \emph{Degenerate clustering} ($K=1$): \mmb collapses to \bpe.
\end{compactitem}
As $K$ grows large, each cluster becomes more specialized, but may also have fewer supporting samples in the validation set, leading to potential overfitting or redundant clusters whose weights remain near uniform. Empirically, we find ($K\approx 16$) obtains a good specialization-stability trade-off.

\xhdr{Extensions.} \mmb readily generalizes beyond TTI quality evaluation. In VQA, it can prioritize prompts whose semantics align with the scene’s content, style, or latent structure, leading to better-calibrated answer probabilities without fine-tuning the base model. The same mechanism can highlight domain-specific cues--such as distinguishing cartoon from photographic violence--in content moderation to reduce false positives while still leveraging closed-source models. For ordinal outputs like Likert ratings, the MMB scaffold remains unchanged; the task’s threshold function can instead return an error interval rather than confidence.

\xhdr{Limitations.}
\label{sec:limitations}
\mmb inherits the typical limitations associated with clustering-based approaches. primarily introducing additional hyperparameters--most notably, the choice of cluster count $K$. Although our experiments indicate that the method is robust to a wide range of values for $K$, excessively large or small values can still impact performance. Additionally, \mmb has greater computational complexity than the original Bayesian Prompt Ensembles due to the embedding of images into clusters and larger number of parameters. However, the computational cost is primarily incurred during the embedding of images in $\mathcal{D}_{sup}$ rather than the optimization and inference of additional prompt parameters. After extracting embeddings, we train and test \mmb on a consumer laptop and have found it only takes 1-2s to train and test our method in a single experiment.

\vspace{-0.1em}
\section{Conclusion}
\label{sec:conclusion}
Assessing multimodal models is an important challenge in LVLM research, especially for text-to-image generation, where subjective factors complicate reliable scoring with automated methods and manual evaluation is costly. Existing MLLM judge models provide a potential alternative but also struggle with inconsistencies, overconfidence, and biases, limiting their usefulness as reliable automated evaluators. This work introduced \mmblong, a novel approach to enhance judge accuracy and calibration by conditioning prompt ensemble weights on clustered image features. Our experiments on HPSv2 and MJBench demonstrate that MMBPE outperforms choosing a single-prompt as well as SOTA ensemble-based methods, achieving stronger calibration and better human alignment in TTI evaluation.

{
    \small
    \bibliographystyle{style/ieeenat_fullname}
    \bibliography{main}

\begin{thebibliography}{74}
\providecommand{\natexlab}[1]{#1}
\providecommand{\url}[1]{\texttt{#1}}
\expandafter\ifx\csname urlstyle\endcsname\relax
  \providecommand{\doi}[1]{doi: #1}\else
  \providecommand{\doi}{doi: \begingroup \urlstyle{rm}\Url}\fi

\bibitem[Anderson et~al.(2016)Anderson, Fernando, Johnson, and Gould]{anderson2016spice}
Peter Anderson, Basura Fernando, Mark Johnson, and Stephen Gould.
\newblock Spice: Semantic propositional image caption evaluation.
\newblock In \emph{Computer Vision--ECCV 2016: 14th European Conference, Amsterdam, The Netherlands, October 11-14, 2016, Proceedings, Part V 14}, pages 382--398. Springer, 2016.

\bibitem[Antol et~al.(2015)Antol, Agrawal, Lu, Mitchell, Batra, Zitnick, and Parikh]{antol2015vqa}
Stanislaw Antol, Aishwarya Agrawal, Jiasen Lu, Margaret Mitchell, Dhruv Batra, C.~Lawrence Zitnick, and Devi Parikh.
\newblock {VQA}: {V}isual {Q}uestion {A}nswering.
\newblock \emph{ICCV}, 2015.

\bibitem[Awadalla et~al.(2023)Awadalla, Gao, Gardner, Hessel, Hanafy, Zhu, Marathe, Bitton, Gadre, Sagawa, Jitsev, Kornblith, Koh, Ilharco, Wortsman, and Schmidt]{awadalla2023openflamingo}
Anas Awadalla, Irena Gao, Josh Gardner, Jack Hessel, Yusuf Hanafy, Wanrong Zhu, Kalyani Marathe, Yonatan Bitton, Samir Gadre, Shiori Sagawa, Jenia Jitsev, Simon Kornblith, Pang~Wei Koh, Gabriel Ilharco, Mitchell Wortsman, and Ludwig Schmidt.
\newblock Openflamingo: An open-source framework for training large autoregressive vision-language models.
\newblock \emph{arXiv}, 2023.

\bibitem[Bai et~al.(2023)Bai, Bai, Yang, Wang, Tan, Wang, Lin, Zhou, and Zhou]{bai2023qwen}
Jinze Bai, Shuai Bai, Shusheng Yang, Shijie Wang, Sinan Tan, Peng Wang, Junyang Lin, Chang Zhou, and Jingren Zhou.
\newblock Qwen-vl: A versatile vision-language model for understanding, localization, text reading, and beyond.
\newblock \emph{arXiv}, 2023.

\bibitem[Benjamini and Yekutieli(2001)]{benjamini2001control}
Yoav Benjamini and Daniel Yekutieli.
\newblock The control of the false discovery rate in multiple testing under dependency.
\newblock \emph{Annals of statistics}, pages 1165--1188, 2001.

\bibitem[Betker et~al.(2023)Betker, Goh, Jing, Brooks, Wang, Li, Ouyang, Zhuang, Lee, Guo, Manassra, Dhariwal, Chu, Jiao, and Ramesh]{betker2023improving}
James Betker, Gabriel Goh, Li Jing, Tim Brooks, Jianfeng Wang, Linjie Li, Long Ouyang, Juntang Zhuang, Joyce Lee, Yufei Guo, Wesam Manassra, Prafulla Dhariwal, Casey Chu, Yunxin Jiao, and Aditya Ramesh.
\newblock Improving image generation with better captions.
\newblock \emph{arXiv}, 2023.

\bibitem[Bitton et~al.(2024)Bitton, Bansal, Hessel, Shao, Zhu, Awadalla, Gardner, Taori, and Schmidt]{bitton2023visitbench}
Yonatan Bitton, Hritik Bansal, Jack Hessel, Rulin Shao, Wanrong Zhu, Anas Awadalla, Josh Gardner, Rohan Taori, and Ludwig Schmidt.
\newblock Visit-bench: A benchmark for vision-language instruction following inspired by real-world use.
\newblock \emph{NeurIPS, Datasets and Benchmarks}, 2024.

\bibitem[Bradley(1997)]{bradley1997auc}
Andrew~P. Bradley.
\newblock The use of the area under the {ROC} curve in the evaluation of machine learning algorithms.
\newblock \emph{Pattern Recognition}, 30\penalty0 (7):\penalty0 1145--1159, 1997.

\bibitem[Brier(1950)]{brier1950score}
Glenn~W. Brier.
\newblock Verification of forecasts expressed in terms of probability.
\newblock \emph{Monthly Weather Review}, 78\penalty0 (1):\penalty0 1--3, 1950.

\bibitem[Chen et~al.(2023)Chen, Liu, and Ding]{chen2023x}
Yixiong Chen, Li Liu, and Chris Ding.
\newblock X-iqe: explainable image quality evaluation for text-to-image generation with visual large language models.
\newblock \emph{arXiv}, 2023.

\bibitem[Chen et~al.(2024)Chen, Du, Wen, Zhou, Cui, Weng, Tu, Wang, Tong, Huang, et~al.]{chen2024mj}
Zhaorun Chen, Yichao Du, Zichen Wen, Yiyang Zhou, Chenhang Cui, Zhenzhen Weng, Haoqin Tu, Chaoqi Wang, Zhengwei Tong, Qinglan Huang, et~al.
\newblock Mj-bench: Is your multimodal reward model really a good judge for text-to-image generation?
\newblock \emph{arXiv preprint arXiv:2407.04842}, 2024.

\bibitem[Cho et~al.(2023)Cho, Zala, and Bansal]{cho2023dalle}
Jaemin Cho, Abhay Zala, and Mohit Bansal.
\newblock Dall-eval: Probing the reasoning skills and social biases of text-to-image generation models.
\newblock \emph{ICCV}, 2023.

\bibitem[Cohen(1960)]{cohen1960kappa}
Jacob Cohen.
\newblock A coefficient of agreement for nominal scales.
\newblock \emph{Educational and Psychological Measurement}, 20\penalty0 (1):\penalty0 37--46, 1960.

\bibitem[Cui et~al.(2024)Cui, Sun, Zhou, and Li]{cui2024exploring}
Xiao Cui, Qi Sun, Wengang Zhou, and Houqiang Li.
\newblock Exploring {GPT}-4 vision for text-to-image synthesis evaluation.
\newblock \emph{ICLR, Tiny Papers}, 2024.

\bibitem[Dai et~al.(2024)Dai, Li, Li, Tiong, Zhao, Wang, Li, Fung, and Hoi]{dai2024instructblip}
Wenliang Dai, Junnan Li, Dongxu Li, Anthony Meng~Huat Tiong, Junqi Zhao, Weisheng Wang, Boyang Li, Pascale~N Fung, and Steven Hoi.
\newblock Instructblip: Towards general-purpose vision-language models with instruction tuning.
\newblock \emph{NeurIPS}, 2024.

\bibitem[Dancette et~al.(2023)Dancette, Whitehead, Maheshwary, Vedantam, Scherer, Chen, Cord, and Rohrbach]{dancette2023improving}
Corentin Dancette, Spencer Whitehead, Rishabh Maheshwary, Ramakrishna Vedantam, Stefan Scherer, Xinlei Chen, Matthieu Cord, and Marcus Rohrbach.
\newblock Improving selective visual question answering by learning from your peers.
\newblock In \emph{Proceedings of the IEEE/CVF Conference on Computer Vision and Pattern Recognition}, pages 24049--24059, 2023.

\bibitem[Das et~al.(2017)Das, Kottur, Gupta, Singh, Yadav, Moura, Parikh, and Batra]{das2017visdial}
Abhishek Das, Satwik Kottur, Khushi Gupta, Avi Singh, Deshraj Yadav, Jos\'e~M.F. Moura, Devi Parikh, and Dhruv Batra.
\newblock {V}isual {D}ialog.
\newblock \emph{CVPR}, 2017.

\bibitem[Davis and Goadrich(2006)]{davis2006relationship}
Jesse Davis and Mark Goadrich.
\newblock The relationship between precision-recall and {ROC} curves.
\newblock In \emph{Proceedings of the 23rd International Conference on Machine Learning (ICML)}, pages 233--240, 2006.

\bibitem[Ge et~al.(2023)Ge, Chen, Chen, Chen, Chen, Yan, Zhu, Lin, Xie, Wang, et~al.]{ge2023mllm}
Wentao Ge, Shunian Chen, Guiming Chen, Junying Chen, Zhihong Chen, Shuo Yan, Chenghao Zhu, Ziyue Lin, Wenya Xie, Xidong Wang, et~al.
\newblock Mllm-bench, evaluating multi-modal llms using gpt-4v.
\newblock \emph{arXiv}, 2023.

\bibitem[Geifman and El-Yaniv(2017)]{geifman2017selective}
Yonatan Geifman and Ran El-Yaniv.
\newblock Selective classification for deep neural networks.
\newblock \emph{NeurIPS}, 30, 2017.

\bibitem[Graves(2011)]{graves2011practical}
Alex Graves.
\newblock Practical variational inference for neural networks.
\newblock \emph{Advances in neural information processing systems}, 24, 2011.

\bibitem[Gu et~al.(2024)Gu, Jiang, Shi, Tan, Zhai, Xu, Li, Shen, Ma, Liu, et~al.]{gu2024survey}
Jiawei Gu, Xuhui Jiang, Zhichao Shi, Hexiang Tan, Xuehao Zhai, Chengjin Xu, Wei Li, Yinghan Shen, Shengjie Ma, Honghao Liu, et~al.
\newblock A survey on llm-as-a-judge.
\newblock \emph{arXiv preprint arXiv:2411.15594}, 2024.

\bibitem[Guo et~al.(2017)Guo, Pleiss, Sun, and Weinberger]{guo2017calibration}
Chuan Guo, Geoff Pleiss, Yu Sun, and Kilian~Q. Weinberger.
\newblock On calibration of modern neural networks.
\newblock In \emph{Proceedings of the 34th International Conference on Machine Learning (ICML)}, pages 1321--1330, 2017.

\bibitem[Hastie et~al.(2001)Hastie, Tibshirani, and Friedman]{hastie2001nll}
Trevor Hastie, Robert Tibshirani, and Jerome Friedman.
\newblock \emph{The Elements of Statistical Learning}.
\newblock Springer, New York, NY, USA, 2001.

\bibitem[He et~al.(2023)He, Zhang, Wang, Kumar, Cho, Glass, and Tsvetkov]{he2022blind}
Tianxing He, Jingyu Zhang, Tianle Wang, Sachin Kumar, Kyunghyun Cho, James~R. Glass, and Yulia Tsvetkov.
\newblock On the blind spots of model-based evaluation metrics for text generation.
\newblock \emph{ACL}, 2023.

\bibitem[Hendrickx et~al.(2024)Hendrickx, Perini, Van~der Plas, Meert, and Davis]{hendrickx2024machine}
Kilian Hendrickx, Lorenzo Perini, Dries Van~der Plas, Wannes Meert, and Jesse Davis.
\newblock Machine learning with a reject option: A survey.
\newblock \emph{Machine Learning}, pages 1--38, 2024.

\bibitem[Heusel et~al.(2017)Heusel, Ramsauer, Unterthiner, Nessler, and Hochreiter]{heusel2017gans}
Martin Heusel, Hubert Ramsauer, Thomas Unterthiner, Bernhard Nessler, and Sepp Hochreiter.
\newblock Gans trained by a two time-scale update rule converge to a local nash equilibrium.
\newblock \emph{Advances in neural information processing systems}, 30, 2017.

\bibitem[Hou et~al.(2023)Hou, O’connor, Andreas, Chang, and Zhang]{hou2023promptboosting}
Bairu Hou, Joe O’connor, Jacob Andreas, Shiyu Chang, and Yang Zhang.
\newblock Promptboosting: Black-box text classification with ten forward passes.
\newblock In \emph{International Conference on Machine Learning}, pages 13309--13324. PMLR, 2023.

\bibitem[Hu et~al.(2024)Hu, Gao, Hu, Zhang, Chen, Xu, and Wan]{hu2024are}
Xinyu Hu, Mingqi Gao, Sen Hu, Yang Zhang, Yicheng Chen, Teng Xu, and Xiaojun Wan.
\newblock Are llm-based evaluators confusing nlg quality criteria?
\newblock \emph{arXiv}, 2024.

\bibitem[Hurst et~al.(2024)Hurst, Lerer, Goucher, Perelman, Ramesh, Clark, Ostrow, Welihinda, Hayes, Radford, et~al.]{hurst2024gpt}
Aaron Hurst, Adam Lerer, Adam~P Goucher, Adam Perelman, Aditya Ramesh, Aidan Clark, AJ Ostrow, Akila Welihinda, Alan Hayes, Alec Radford, et~al.
\newblock Gpt-4o system card.
\newblock \emph{arXiv preprint arXiv:2410.21276}, 2024.

\bibitem[Jiang et~al.(2023)Jiang, Ruan, Huang, Liao, Pitis, Grosse, and Ba]{jiang2023calibrating}
Mingjian Jiang, Yangjun Ruan, Sicong Huang, Saifei Liao, Silviu Pitis, Roger~Baker Grosse, and Jimmy Ba.
\newblock Calibrating language models via augmented prompt ensembles.
\newblock \emph{ICML Workshop on Deployable Generative AI}, 2023.

\bibitem[Jiang et~al.(2020)Jiang, Xu, Araki, and Neubig]{jiang2020can}
Zhengbao Jiang, Frank~F Xu, Jun Araki, and Graham Neubig.
\newblock How can we know what language models know?
\newblock \emph{Transactions of the Association for Computational Linguistics}, 8:\penalty0 423--438, 2020.

\bibitem[Johnson et~al.(2019)Johnson, Douze, and J{\'e}gou]{johnson2019faiss}
Jeff Johnson, Matthijs Douze, and Herv{\'e} J{\'e}gou.
\newblock Billion-scale similarity search with {GPUs}.
\newblock \emph{IEEE Transactions on Big Data}, 2019.

\bibitem[Kirstain et~al.(2023)Kirstain, Polyak, Singer, Matiana, Penna, and Levy]{kirstain2023pick}
Yuval Kirstain, Adam Polyak, Uriel Singer, Shahbuland Matiana, Joe Penna, and Omer Levy.
\newblock Pick-a-pic: An open dataset of user preferences for text-to-image generation.
\newblock \emph{arXiv preprint arXiv:2305.01569}, 2023.

\bibitem[Koo et~al.(2023)Koo, Lee, Raheja, Park, Kim, and Kang]{koo2023benchmarking}
Ryan Koo, Minhwa Lee, Vipul Raheja, Jong~Inn Park, Zae~Myung Kim, and Dongyeop Kang.
\newblock Benchmarking cognitive biases in large language models as evaluators.
\newblock \emph{arXiv}, 2023.

\bibitem[Kynk{\"a}{\"a}nniemi et~al.(2019)Kynk{\"a}{\"a}nniemi, Karras, Laine, Lehtinen, and Aila]{kynkaanniemi2019improved}
Tuomas Kynk{\"a}{\"a}nniemi, Tero Karras, Samuli Laine, Jaakko Lehtinen, and Timo Aila.
\newblock Improved precision and recall metric for assessing generative models.
\newblock \emph{Advances in neural information processing systems}, 32, 2019.

\bibitem[Li et~al.(2024{\natexlab{a}})Li, Ge, Ge, Wang, Wang, Zhang, and Shan]{li2023seed}
Bohao Li, Yuying Ge, Yixiao Ge, Guangzhi Wang, Rui Wang, Ruimao Zhang, and Ying Shan.
\newblock Seed-bench: Benchmarking multimodal large language models.
\newblock \emph{CVPR}, 2024{\natexlab{a}}.

\bibitem[Li et~al.(2024{\natexlab{b}})Li, Jiang, Huang, Beigi, Zhao, Tan, Bhattacharjee, Jiang, Chen, Wu, Shu, Cheng, and Liu]{li2024llmasajudge}
Dawei Li, Bohan Jiang, Liangjie Huang, Alimohammad Beigi, Chengshuai Zhao, Zhen Tan, Amrita Bhattacharjee, Yuxuan Jiang, Canyu Chen, Tianhao Wu, Kai Shu, Lu Cheng, and Huan Liu.
\newblock From generation to judgment: Opportunities and challenges of llm-as-a-judge.
\newblock \emph{arXiv preprint arXiv: 2411.16594}, 2024{\natexlab{b}}.

\bibitem[Li et~al.(2024{\natexlab{c}})Li, Dong, Chen, Su, Zhou, Ai, Ye, and Liu]{li2024llms}
Haitao Li, Qian Dong, Junjie Chen, Huixue Su, Yujia Zhou, Qingyao Ai, Ziyi Ye, and Yiqun Liu.
\newblock Llms-as-judges: a comprehensive survey on llm-based evaluation methods.
\newblock \emph{arXiv preprint arXiv:2412.05579}, 2024{\natexlab{c}}.

\bibitem[Lin(2004)]{lin2004rouge}
Chin-Yew Lin.
\newblock {ROUGE}: A package for automatic evaluation of summaries.
\newblock \emph{Text Summarization Branches Out}, 2004.

\bibitem[Lin et~al.(2014)Lin, Maire, Belongie, Hays, Perona, Ramanan, Doll{\'a}r, and Zitnick]{lin2014coco}
Tsung-Yi Lin, Michael Maire, Serge Belongie, James Hays, Pietro Perona, Deva Ramanan, Piotr Doll{\'a}r, and C~Lawrence Zitnick.
\newblock Microsoft coco: Common objects in context.
\newblock \emph{ECCV}, 2014.

\bibitem[Liu et~al.(2024)Liu, Lin, Li, Wang, Yacoob, and Wang]{liu2024mitigating}
Fuxiao Liu, Kevin Lin, Linjie Li, Jianfeng Wang, Yaser Yacoob, and Lijuan Wang.
\newblock Mitigating hallucination in large multi-modal models via robust instruction tuning.
\newblock \emph{ICLR}, 2024.

\bibitem[Liu et~al.(2023{\natexlab{a}})Liu, Li, Wu, and Lee]{liu2023llava}
Haotian Liu, Chunyuan Li, Qingyang Wu, and Yong~Jae Lee.
\newblock Visual instruction tuning.
\newblock \emph{NeurIPS}, 2023{\natexlab{a}}.

\bibitem[Liu et~al.(2023{\natexlab{b}})Liu, Moosavi, and Lin]{liu2023llms}
Yiqi Liu, Nafise~Sadat Moosavi, and Chenghua Lin.
\newblock Llms as narcissistic evaluators: When ego inflates evaluation scores.
\newblock \emph{arXiv}, 2023{\natexlab{b}}.

\bibitem[Madras et~al.(2018)Madras, Pitassi, and Zemel]{madras2018predict}
David Madras, Toni Pitassi, and Richard Zemel.
\newblock Predict responsibly: improving fairness and accuracy by learning to defer.
\newblock \emph{NeurIPS}, 31, 2018.

\bibitem[OpenAI(2023)]{openai2023gpt4v}
OpenAI.
\newblock Gpt-4v(ision) technical work and authors.
\newblock \url{https://openai.com/contributions/gpt-4v/}, 2023.

\bibitem[OpenAI(2024)]{openai2025o1}
OpenAI.
\newblock Openai o1 system card.
\newblock \url{https://openai.com/index/openai-o1-system-card/}, 2024.

\bibitem[Panickssery et~al.(2024)Panickssery, Bowman, and Feng]{panickssery2024llm}
Arjun Panickssery, Samuel~R. Bowman, and Shi Feng.
\newblock Llm evaluators recognize and favor their own generations.
\newblock \emph{arXiv}, 2024.

\bibitem[Papineni et~al.(2002)Papineni, Roukos, Ward, and Zhu]{papineni2002bleu}
Kishore Papineni, Salim Roukos, Todd Ward, and Wei-Jing Zhu.
\newblock Bleu: a method for automatic evaluation of machine translation.
\newblock \emph{ACL}, 2002.

\bibitem[Peng et~al.(2023)Peng, Wang, Dong, Hao, Huang, Ma, and Wei]{peng2023kosmos}
Zhiliang Peng, Wenhui Wang, Li Dong, Yaru Hao, Shaohan Huang, Shuming Ma, and Furu Wei.
\newblock Kosmos-2: Grounding multimodal large language models to the world.
\newblock \emph{arXiv}, 2023.

\bibitem[Podell et~al.(2024)Podell, English, Lacey, Blattmann, Dockhorn, M{\"u}ller, Penna, and Rombach]{podell2023sdxl}
Dustin Podell, Zion English, Kyle Lacey, Andreas Blattmann, Tim Dockhorn, Jonas M{\"u}ller, Joe Penna, and Robin Rombach.
\newblock Sdxl: improving latent diffusion models for high-resolution image synthesis.
\newblock In \emph{International Conference on Learning Representations}, pages 1--13, 2024.

\bibitem[Radford et~al.(2021{\natexlab{a}})Radford, Kim, Hallacy, Ramesh, Goh, Agarwal, Sastry, Askell, Mishkin, Clark, Krueger, and Sutskever]{radford2021clip}
Alec Radford, Jong~Wook Kim, Chris Hallacy, A. Ramesh, Gabriel Goh, Sandhini Agarwal, Girish Sastry, Amanda Askell, Pamela Mishkin, Jack Clark, Gretchen Krueger, and Ilya Sutskever.
\newblock Learning transferable visual models from natural language supervision.
\newblock \emph{ICML}, 2021{\natexlab{a}}.

\bibitem[Radford et~al.(2021{\natexlab{b}})Radford, Kim, Hallacy, Ramesh, Goh, Agarwal, Sastry, Askell, Mishkin, Clark, et~al.]{radford2021learning}
Alec Radford, Jong~Wook Kim, Chris Hallacy, Aditya Ramesh, Gabriel Goh, Sandhini Agarwal, Girish Sastry, Amanda Askell, Pamela Mishkin, Jack Clark, et~al.
\newblock Learning transferable visual models from natural language supervision.
\newblock In \emph{International conference on machine learning}, pages 8748--8763. PMLR, 2021{\natexlab{b}}.

\bibitem[Rombach et~al.(2022)Rombach, Blattmann, Lorenz, Esser, and Ommer]{rombach2022high}
Robin Rombach, Andreas Blattmann, Dominik Lorenz, Patrick Esser, and Bj{\"o}rn Ommer.
\newblock High-resolution image synthesis with latent diffusion models.
\newblock In \emph{Proceedings of the IEEE/CVF conference on computer vision and pattern recognition}, pages 10684--10695, 2022.

\bibitem[Saharia et~al.(2022)Saharia, Chan, Saxena, Li, Whang, Denton, Ghasemipour, Gontijo~Lopes, Karagol~Ayan, Salimans, et~al.]{saharia2022photorealistic}
Chitwan Saharia, William Chan, Saurabh Saxena, Lala Li, Jay Whang, Emily~L Denton, Kamyar Ghasemipour, Raphael Gontijo~Lopes, Burcu Karagol~Ayan, Tim Salimans, et~al.
\newblock Photorealistic text-to-image diffusion models with deep language understanding.
\newblock \emph{Advances in neural information processing systems}, 35:\penalty0 36479--36494, 2022.

\bibitem[Saito et~al.(2023)Saito, Wachi, Wataoka, and Akimoto]{saito2023verbosity}
Keita Saito, Akifumi Wachi, Koki Wataoka, and Youhei Akimoto.
\newblock Verbosity bias in preference labeling by large language models.
\newblock \emph{arXiv preprint arXiv:2310.10076}, 2023.

\bibitem[Salimans et~al.(2016)Salimans, Goodfellow, Zaremba, Cheung, Radford, and Chen]{salimans2016improved}
Tim Salimans, Ian Goodfellow, Wojciech Zaremba, Vicki Cheung, Alec Radford, and Xi Chen.
\newblock Improved techniques for training gans.
\newblock \emph{Advances in neural information processing systems}, 29, 2016.

\bibitem[Team et~al.(2023)Team, Anil, Borgeaud, Alayrac, Yu, Soricut, Schalkwyk, Dai, Hauth, Millican, et~al.]{team2023gemini}
Gemini Team, Rohan Anil, Sebastian Borgeaud, Jean-Baptiste Alayrac, Jiahui Yu, Radu Soricut, Johan Schalkwyk, Andrew~M Dai, Anja Hauth, Katie Millican, et~al.
\newblock Gemini: a family of highly capable multimodal models.
\newblock \emph{arXiv preprint arXiv:2312.11805}, 2023.

\bibitem[Tonolini et~al.(2024)Tonolini, Aletras, Massiah, and Kazai]{tonolini2024bayesian}
Francesco Tonolini, Nikolaos Aletras, Jordan Massiah, and Gabriella Kazai.
\newblock Bayesian prompt ensembles: Model uncertainty estimation for black-box large language models.
\newblock In \emph{Findings of the Association for Computational Linguistics ACL 2024}, pages 12229--12272, 2024.

\bibitem[Wang et~al.(2023)Wang, Li, Chen, Zhu, Lin, Cao, Liu, Liu, and Sui]{wang2023large}
Peiyi Wang, Lei Li, Liang Chen, Dawei Zhu, Binghuai Lin, Yunbo Cao, Qi Liu, Tianyu Liu, and Zhifang Sui.
\newblock Large language models are not fair evaluators.
\newblock \emph{arXiv}, 2023.

\bibitem[White et~al.(2023)White, Fu, Hays, Sandborn, Olea, Gilbert, Elnashar, Spencer-Smith, and Schmidt]{white2023prompt}
Jules White, Quchen Fu, Sam Hays, Michael Sandborn, Carlos Olea, Henry Gilbert, Ashraf Elnashar, Jesse Spencer-Smith, and Douglas~C Schmidt.
\newblock A prompt pattern catalog to enhance prompt engineering with chatgpt.
\newblock \emph{arXiv preprint arXiv:2302.11382}, 2023.

\bibitem[Whitehead et~al.(2022)Whitehead, Petryk, Shakib, Gonzalez, Darrell, Rohrbach, and Rohrbach]{whitehead2022reliable}
Spencer Whitehead, Suzanne Petryk, Vedaad Shakib, Joseph Gonzalez, Trevor Darrell, Anna Rohrbach, and Marcus Rohrbach.
\newblock Reliable visual question answering: Abstain rather than answer incorrectly.
\newblock In \emph{European Conference on Computer Vision}, pages 148--166. Springer, 2022.

\bibitem[Wightman et~al.(2023)Wightman, Delucia, and Dredze]{wightman2023strength}
Gwenyth~Portillo Wightman, Alexandra Delucia, and Mark Dredze.
\newblock Strength in numbers: Estimating confidence of large language models by prompt agreement.
\newblock In \emph{Proceedings of the 3rd Workshop on Trustworthy Natural Language Processing (TrustNLP 2023)}, pages 326--362, 2023.

\bibitem[Wu et~al.(2024)Wu, Yang, Li, Zhang, Liu, Guibas, Lin, and Wetzstein]{wu2023gpteval3d}
Tong Wu, Guandao Yang, Zhibing Li, Kai Zhang, Ziwei Liu, Leonidas Guibas, Dahua Lin, and Gordon Wetzstein.
\newblock Gpt-4v(ision) is a human-aligned evaluator for text-to-3d generation.
\newblock \emph{CVPR}, 2024.

\bibitem[Wu et~al.(2023)Wu, Hao, Sun, Chen, Zhu, Zhao, and Li]{wu2023human}
Xiaoshi Wu, Yiming Hao, Keqiang Sun, Yixiong Chen, Feng Zhu, Rui Zhao, and Hongsheng Li.
\newblock Human preference score v2: A solid benchmark for evaluating human preferences of text-to-image synthesis.
\newblock \emph{arXiv preprint arXiv:2306.09341}, 2023.

\bibitem[Xiong et~al.(2024)Xiong, Wang, Guo, Ye, Fan, Gu, Huang, and Li]{xiong2024llava}
Tianyi Xiong, Xiyao Wang, Dong Guo, Qinghao Ye, Haoqi Fan, Quanquan Gu, Heng Huang, and Chunyuan Li.
\newblock Llava-critic: Learning to evaluate multimodal models.
\newblock \emph{arXiv preprint arXiv:2410.02712}, 2024.

\bibitem[Yan et~al.(2024)Yan, Bai, Chen, Zhou, Huang, and Li]{yan2024vigor}
Siming Yan, Min Bai, Weifeng Chen, Xiong Zhou, Qixing Huang, and Li~Erran Li.
\newblock Vigor: Improving visual grounding of large vision language models with fine-grained reward modeling.
\newblock \emph{arXiv}, 2024.

\bibitem[Ye et~al.(2024)Ye, Wang, Huang, Chen, Zhang, Moniz, Gao, Geyer, Huang, Chen, Chawla, and Zhang]{ye2024justice}
Jiayi Ye, Yanbo Wang, Yue Huang, Dongping Chen, Qihui Zhang, Nuno Moniz, Tian Gao, Werner Geyer, Chao Huang, Pin-Yu Chen, Nitesh~V Chawla, and Xiangliang Zhang.
\newblock Justice or prejudice? quantifying biases in llm-as-a-judge.
\newblock \emph{arXiv preprint arXiv:2410.02736}, 2024.

\bibitem[Yin et~al.(2023)Yin, Wang, Cao, Shi, Liu, Li, Sheng, Bai, Huang, Wang, et~al.]{yin2023lamm}
Zhenfei Yin, Jiong Wang, Jianjian Cao, Zhelun Shi, Dingning Liu, Mukai Li, Lu Sheng, Lei Bai, Xiaoshui Huang, Zhiyong Wang, et~al.
\newblock Lamm: Language-assisted multi-modal instruction-tuning dataset, framework, and benchmark.
\newblock \emph{NeurIPS, Datasets and Benchmarks}, 2023.

\bibitem[Young et~al.(2014)Young, Lai, Hodosh, and Hockenmaier]{young2014flickr}
Peter Young, Alice Lai, Micah Hodosh, and Julia Hockenmaier.
\newblock From image descriptions to visual denotations: New similarity metrics for semantic inference over event descriptions.
\newblock \emph{TACL}, 2014.

\bibitem[Zhang et~al.(2024)Zhang, Han, Liu, Gao, Zhou, Hu, Yan, Lu, Li, and Qiao]{zhang2024llama}
Renrui Zhang, Jiaming Han, Chris Liu, Peng Gao, Aojun Zhou, Xiangfei Hu, Shilin Yan, Pan Lu, Hongsheng Li, and Yu Qiao.
\newblock Llama-adapter: Efficient fine-tuning of language models with zero-init attention.
\newblock \emph{ICLR}, 2024.

\bibitem[Zhang et~al.(2023)Zhang, Lu, Wang, Yan, Yan, Qin, Wang, Yan, Wang, and Petzold]{zhang2023gpt}
Xinlu Zhang, Yujie Lu, Weizhi Wang, An Yan, Jun Yan, Lianke Qin, Heng Wang, Xifeng Yan, William~Yang Wang, and Linda~Ruth Petzold.
\newblock Gpt-4v (ision) as a generalist evaluator for vision-language tasks.
\newblock \emph{arXiv}, 2023.

\bibitem[Zheng et~al.(2023)Zheng, Chiang, Sheng, Zhuang, Wu, Zhuang, Lin, Li, Li, Xing, Zhang, Gonzalez, and Stoica]{zheng2023judging}
Lianmin Zheng, Wei-Lin Chiang, Ying Sheng, Siyuan Zhuang, Zhanghao Wu, Yonghao Zhuang, Zi Lin, Zhuohan Li, Dacheng Li, Eric~P. Xing, Haotong Zhang, Joseph Gonzalez, and Ion Stoica.
\newblock Judging llm-as-a-judge with mt-bench and chatbot arena.
\newblock \emph{NeurIPS}, 2023.

\bibitem[Zhu et~al.(2024)Zhu, Chen, Shen, Li, and Elhoseiny]{zhu2024minigpt}
Deyao Zhu, Jun Chen, Xiaoqian Shen, Xiang Li, and Mohamed Elhoseiny.
\newblock Minigpt-4: Enhancing vision-language understanding with advanced large language models.
\newblock \emph{ICLR}, 2024.

\end{thebibliography}
}

\clearpage
\setcounter{page}{1}
\maketitlesupplementary
\appendix

\section{Additional Experimental Details}
\label{sec:suppliment}

\begin{table}[ht!]
\centering
\footnotesize
\begin{tabular}{llll}
\toprule
& \textbf{Factor} & \textbf{Levels/Value}  & \textbf{Notes} \\ 
\midrule
\multirow{3}{*}{\rcnt{\scriptsize Exps.}}
& prompts         & 5, 10, 20        & Num. prompts in $\boldsymbol{a}$ \\
& samples         & 5, 10, 20, 50    & Num. samples in $\Dval$ \\
& clusters        & 4, 8, 16, 32, 64 & Num. clusters $K$ \\
\arrayrulecolor{black!20}\midrule
\multirow{3}{*}{\rcnt{\scriptsize Seeds}} 
& train        &  3 Unique & Seed for training \\
& data         & 50 Unique & Seed for sampling $\Dval$ \\
& cluster      &  5 Unique & Seed for sampling $\Dsup$ \\ 
\midrule
\multirow{4}{*}{\rcnt{\scriptsize Clustering}} 
& method  & Spherical & KMeans version \\ 
& samples &  $256\times K$ & Num. clustering samples \\
& init    & 3 & Num. random inits \\
& niter   & 1,000 & Num. training iterations \\
\midrule
\multirow{5}{*}{\rcnt{\scriptsize Optimization}} 
& Optim & L-BFGS & Optimizer \\
& lr & 0.01 &  Learning rate \\
& history & 50 &  History size \\
& max iter & 100 &  Max iterations \\
& search & strong wolfe & Search func \\
\midrule
\multirow{2}{*}{\rcnt{\scriptsize $\phi_I$}} 
& Model & CLIP-ViT-B16 & \cite{radford2021clip} \\
& Weights & {\small laion2b\_s34b\_b88k} & \\
\arrayrulecolor{black}
\bottomrule
\end{tabular}
\caption{Summary of experimental configurations..}
\label{tab:small-experiments}
\end{table}

See table \cref{tab:small-experiments} for details on experimental factors, clustering configuration, and other hyperparameters.

\newpage

\section{Generating Instruction Prompts}
\label{sec:supp-prompts}

We employ a variety of methods to contruct our prompt set $\boldsymbol{a}.$ We both manually, and with the aid of GPT, construct lists of personas, prompt templates, and task instruction criteria. We also take these original templates and create \myquote{augmented} versions by flipping the order of inputs in the template and changing the response glyph (\eg \myquote{A)} \vs \myquote{1.}). We provide a sample prompt for reference ---

\begin{Verbatim}[breaklines=true]
You are a technical expert at evaluating 'text-to-image' alignment and aesthetics. Your task is to assess the quality of two images generated from the same prompt. The criteria for evaluation are as follows:

Image Quality - The image should have a well-balanced composition with clear framing and focal point, effective brightness and contrast in lighting, appealing color harmony and saturation, sharp focus with visible fine details and minimal digital noise, and be high resolution without pixelation.

Image Artifacts - The image should not have any obvious artifacts including excessive blur, occlusion, warping, or other issues.

Be objective in your evaluation, do not consider attributes like age, race, gender, or other demographic information. Respond with a single letter only.

[IMAGE 1]
[IMAGE 2]
Caption: [INPUT PROMPT]

Which of the two images do you prefer?
A) I prefer the first image.
B) I prefer the second image.
\end{Verbatim}

\clearpage
\onecolumn
\section{Derivation of \mmblong}
\label{sec:supp-derivation}
\begin{eqnarray*}
    \log p(y|x) &=& \log \sum_z\sum_a p(y,a,z|x)\\
    &=& \log \sum_z\sum_ap(y|x,a)p(a|z)p(z|x)\\
     &=& \log \sum_z\sum_a p(y|x,a)p(a|z)p(z|x) * \frac{q(a |z)}{q(a | z)}\\
     &=& \log E_{a\sim q(a|z), z\sim p(z|x)} \left[ p(y|x,a) \frac{p(a|z)}{q(a|z)} \right]\\
      &\geq& E_{a\sim q(a|z), z\sim p(z|x)} \left[ \log p(y|x,a) + \log \frac{p(a|z)}{q(a|z)} \right]\\
     &=& E_{a\sim q(a|z), z\sim p(z|x)} \left[\log p(y|x,a)\right] - E_{a\sim q(a|z), z\sim p(z|x)} \left[ \log \frac{q(a|z)}{p(a|z)}\right]\\
     &=& \left[\sum_z p(z|x) \sum_a q(a|z) \log p(y|x,a) \right] - \left[\sum_z p(z|x) KL(q(a|z) || p(a|z)) \right]\\
     &=& \sum_z p(z|x) \left[\left[\sum_a q(a|z) \log p(y|x,a) \right] -  KL\left(q(a|z) || p(a|z)\right)\right]
\end{eqnarray*}

\noindent Assuming uniform priors for $p(a{\mid}z)$ and parameterizing $q(a|z)=w_{za}$, we can write the training objective for \mmblong to maximize this lower-bound on the log-likelihood of all $M$ datapoints in $\Dval$ as:
\begin{eqnarray}
    \operatorname*{arg\,max}_{\bf w}  \sum_{j=1}^M \sum_z p(z|x_j) \left[\sum_a w_{za} \log p(y_j^*|x_j,a) - \sum_a w_{za} \log w_{za}~~ \right]
\end{eqnarray}

\clearpage
\section{Additional Experiment Configurations}
\label{sec:supp-additional-results}
\begin{table*}[!h]
\centering
\footnotesize
\begin{tabular}{ll@{}cccccccccccccccccc}
\toprule
\multirow{2}{*}[-2pt]{\rotatebox{50}{\scriptsize prompts}} & 
\multirow{2}{*}[-2pt]{\rotatebox{50}{\scriptsize samples}} & 
\multicolumn{5}{c}{Expected Calibration Error ($\downarrow$)} & 
\multicolumn{5}{c}{Max Calibration Error ($\downarrow$)} &
\multicolumn{5}{c}{AUC Precision-Recall ($\uparrow$)} \\
\cmidrule(l{3pt}r{3pt}){3-7}
\cmidrule(l{3pt}r{3pt}){8-12}
\cmidrule(l{3pt}r{3pt}){13-17}
& 
& 
\makecell{\textbf{4}} &
\makecell{\textbf{8}} & 
\makecell{\textbf{16}} & 
\makecell{\textbf{32}} & 
\makecell{\textbf{64}} & 
\makecell{\textbf{4}} &
\makecell{\textbf{8}} & 
\makecell{\textbf{16}} & 
\makecell{\textbf{32}} & 
\makecell{\textbf{64}} & 
\makecell{\textbf{4}} &
\makecell{\textbf{8}} & 
\makecell{\textbf{16}} & 
\makecell{\textbf{32}} & 
\makecell{\textbf{64}} \\
\midrule
\ \   5  & \ \  5   &   \textbf{.113} & \textbf{.113} & \textbf{.113} & \textbf{.113} & \textbf{.113} & \textbf{.245} & \textbf{.244} & \textbf{.245} & \textbf{.245} & \textbf{.246} & .834  &  .835  &  .835  &  .835  & \textbf{.836} \\
         &     10   &   \textbf{.107} &  .107  &  .108  &  .108  & \textbf{.108} & \textbf{.239} & \textbf{.239} & \textbf{.239} & \textbf{.239} & \textbf{.239} & .837  &  .837  &  .838  &  .838  & \textbf{.838} \\
         &     20   &   \textbf{.108} &  .108  &  .108  &  .108  & \textbf{.109} & \textbf{.241} & \textbf{.240} & \textbf{.241} &  .241  & \textbf{.242} & .837  &  .837  &  .838  &  .838  & \textbf{.838} \\
         &     50   &   \textbf{.107} & \textbf{.107} &  .107  &  .107  & \textbf{.107} & \textbf{.235} & \textbf{.235} &  .236  &  .236  & \textbf{.237} & .839  &  .839  &  .839  &  .839  & \textbf{.839} \\
\arrayrulecolor{black!20}\midrule
10       & \ \  5   &   \textbf{.092} &  .092  &  .093  &  .092  & \textbf{.093} & \textbf{.199} & \textbf{.200} &  .201  &  .201  & \textbf{.202} & .842  &  .842  &  .843  &  .844  & \textbf{.844}  \\
         &     10   &   \textbf{.094} & \textbf{.094} &  .095  &  .095  & \textbf{.095} & \textbf{.194} & \textbf{.195} &  .196  &  .196  & \textbf{.197} & .843  &  .843  &  .844  &  .844  & \textbf{.845}  \\
         &     20   &   \textbf{.090} & \textbf{.090} &  .091  &  .091  & \textbf{.091} & \textbf{.188} & \textbf{.188} &  .189  &  .189  & \textbf{.189} & .844  &  .844  &  .845  &  .845  & \textbf{.845}  \\
         &     50   &   \textbf{.088} & \textbf{.088} &  .088  &  .088  & \textbf{.089} & \textbf{.187} & \textbf{.187} &  .188  &  .188  & \textbf{.188} & .845  &  .845  &  .845  &  .845  & \textbf{.845}  \\
\midrule
20       & \ \  5   &   \textbf{.078} &  .079  &  .080  &  .080  & \textbf{.080} & \textbf{.170} &  .172  &  .172  &  .172  & \textbf{.173} & .845  &  .846  &  .847  &  .847  & \textbf{.848}  \\
         &     10   &  \textbf{.081} & \textbf{.081} &  .082  &  .082  & \textbf{.082} & \textbf{.168} & \textbf{.169} & \textbf{.169} & \textbf{.169} & \textbf{.169} & .846  &  .847  &  .847  &  .848  & \textbf{.848}  \\
         &     20   &   \textbf{.080} & \textbf{.080} &  .080  &  .080  & \textbf{.081} & \textbf{.159} & \textbf{.159} & \textbf{.159} & \textbf{.160} & \textbf{.161} & .847  &  .847  &  .848  &  .848  & \textbf{.848}  \\
         &     50   &   \textbf{.076} &  .076  &  .076  &  .076  & \textbf{.077} & \textbf{.153} & \textbf{.153} & \textbf{.153} & \textbf{.154} & \textbf{.154} & .848  &  .848  &  .849  &  .849  & \textbf{.849}  \\
\arrayrulecolor{black}\bottomrule
\end{tabular}
\caption{Expected calibration error. Lower is better. Multiple cluster sizes. FDR controlled with Benjamini-Yekutieli~\citep{benjamini2001control}. HPSv2.}
\label{tab:big-res-cluster}
\end{table*}

\begin{figure*}[!h]
    \centering
    \includegraphics[width=0.8\textwidth]{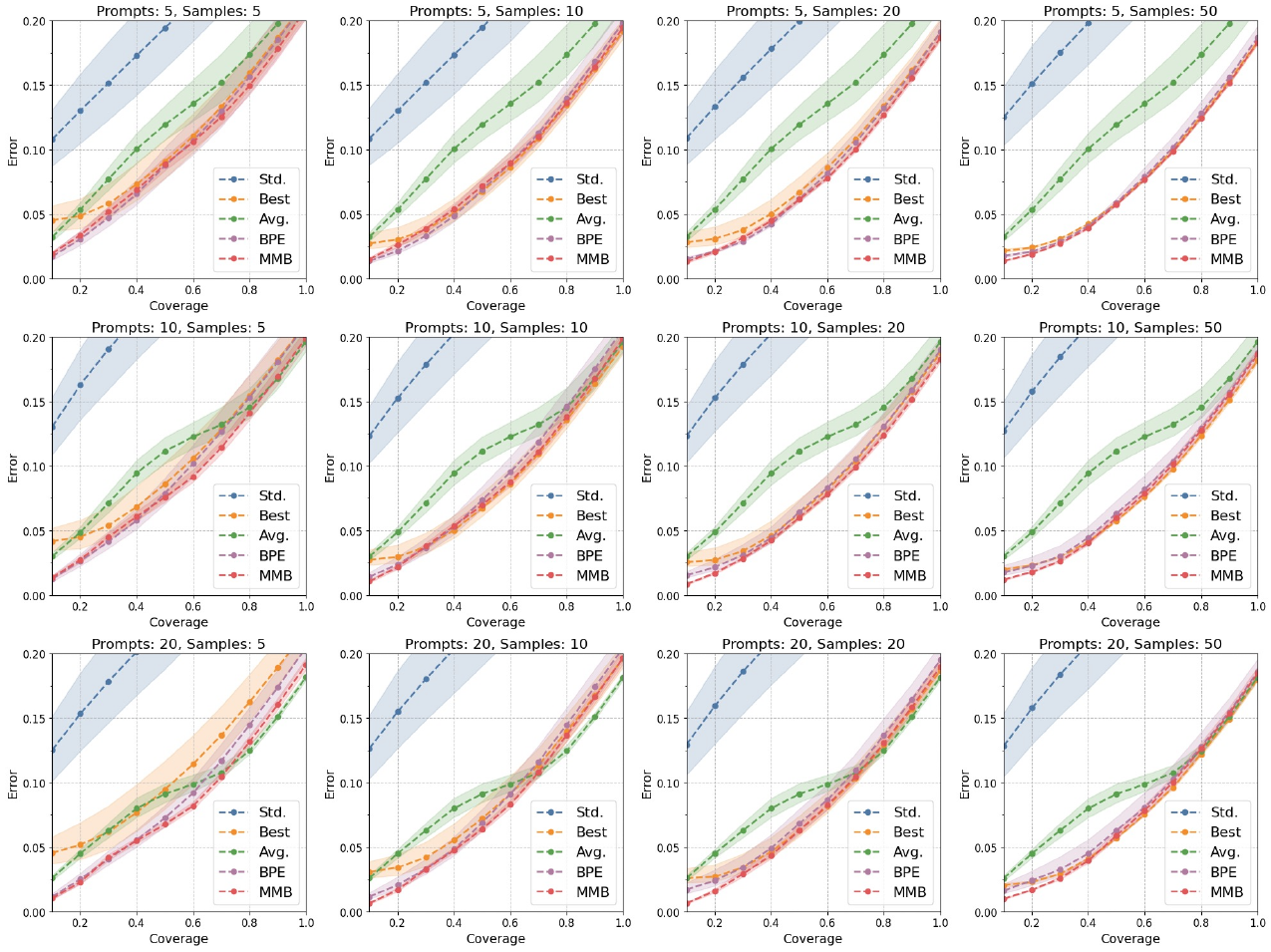}
    \caption{Error-coverage curves for different models. HPSv2.}
    \label{fig:big-coverage}
\end{figure*}

\clearpage
\section{Additional Qualitative Examples}
\label{sec:supp-qualitative}
\begin{figure*}[!h]
    \centering

    \begin{subfigure}[b]{0.80\textwidth}
        \centering
        \includegraphics[width=\textwidth]{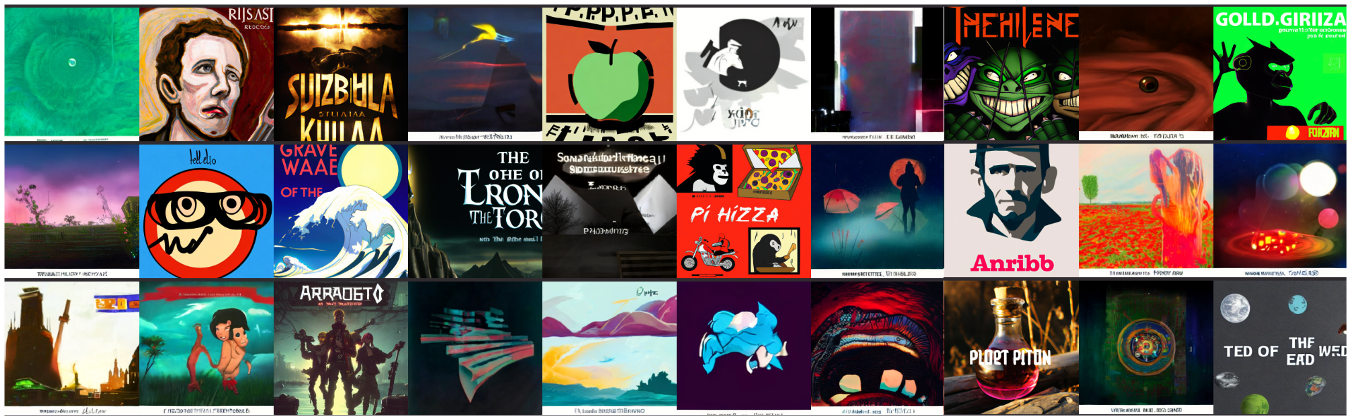}
        \caption{A non-cohesive cluster that results in near-average weights across prompts due to low-validation sample match.}
        \label{fig:big-qualitative-1-1}
    \end{subfigure}

    \begin{subfigure}[b]{0.80\textwidth}
        \centering
        \includegraphics[width=\textwidth]{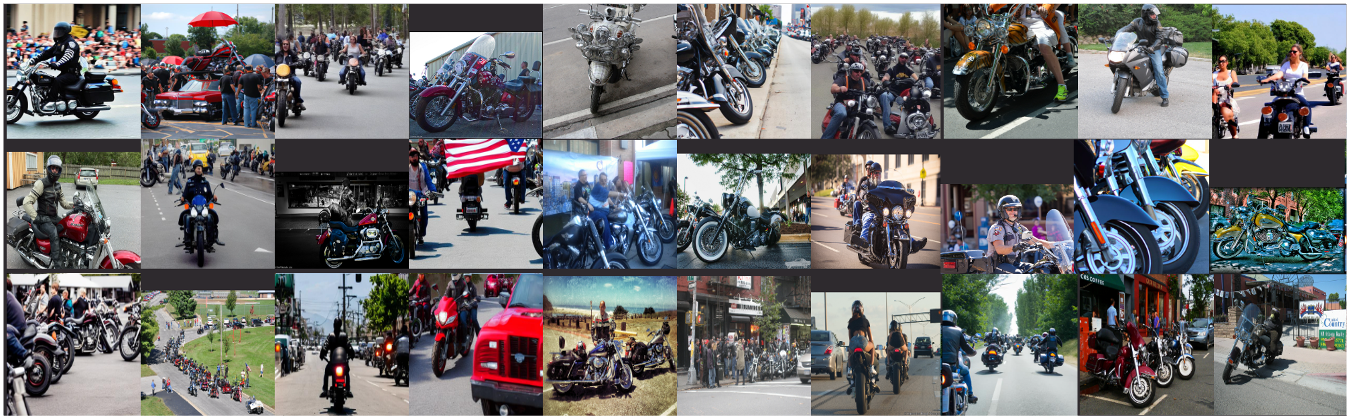}
        \caption{A cohesive cluster which can be matched with validation samples, but does not have any highly weighted prompts.}
        \label{fig:big-qualitative-1-2}
    \end{subfigure}

    % Fourth row - images 7-8
    \begin{subfigure}[b]{0.80\textwidth}
        \centering
        \includegraphics[width=\textwidth]{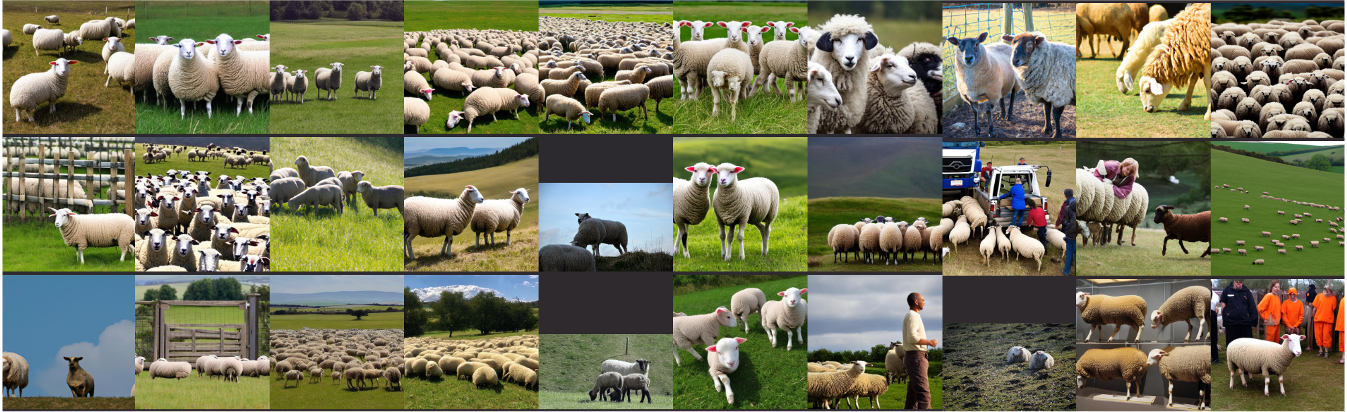}
        \caption{\myquote{You are a \textbf{photographer} skilled in assessing lighting, focus, and overall image sharpness [...]}}
        \label{fig:big-qualitative-1-3}
    \end{subfigure}
    
    \begin{subfigure}[b]{0.80\textwidth}
        \centering
        \includegraphics[width=\textwidth]{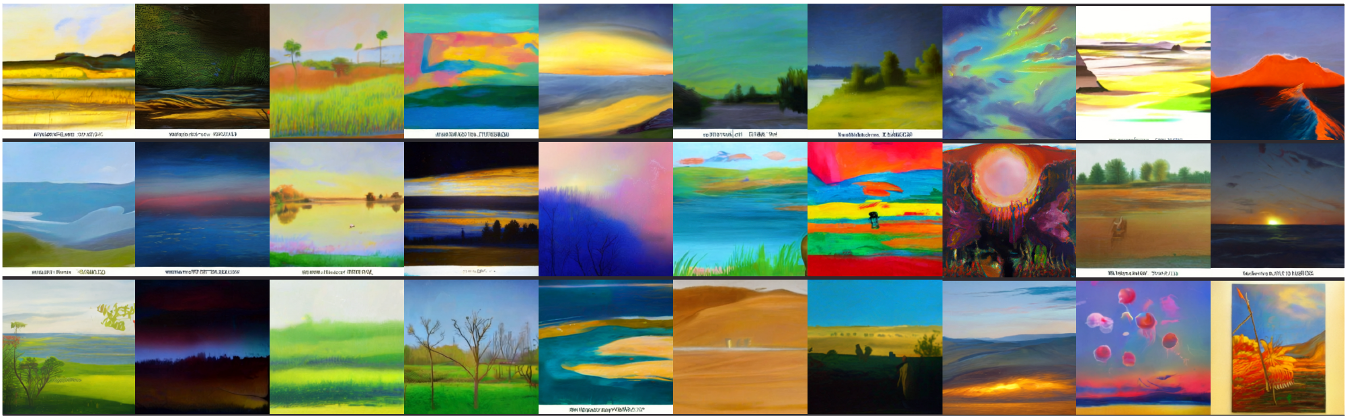}
        \caption{\myquote{You are a \textbf{landscape artist} skilled in assessing lighting, color, and composition [...]}}
        \label{fig:big-qualitative-1-4}
    \end{subfigure}

    \caption{Image clusters and their corresponding highest weighted prompts (or lack thereof) when using $K{=}64$, $N{=}200$.}
    \label{fig:big-qualitative-1}
\end{figure*}

\begin{figure*}[h!]
    \centering

    \begin{subfigure}[b]{0.80\textwidth}
        \centering
        \includegraphics[width=\textwidth]{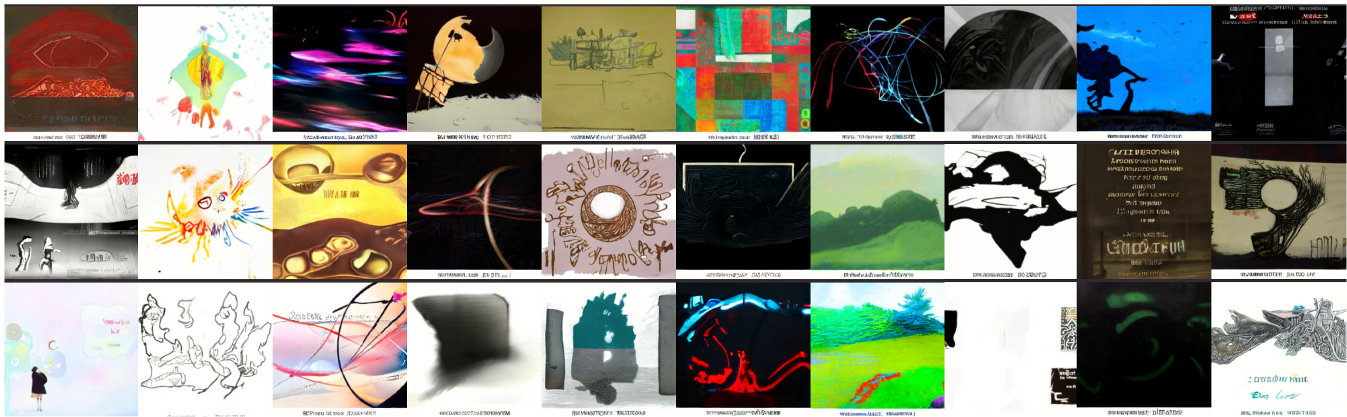}
        \caption{A non-cohesive cluster that results in near-average weights across prompts due to low-validation sample match.}
        \label{fig:big-qualitative-2-1}
    \end{subfigure}

    \begin{subfigure}[b]{0.80\textwidth}
        \centering
        \includegraphics[width=\textwidth]{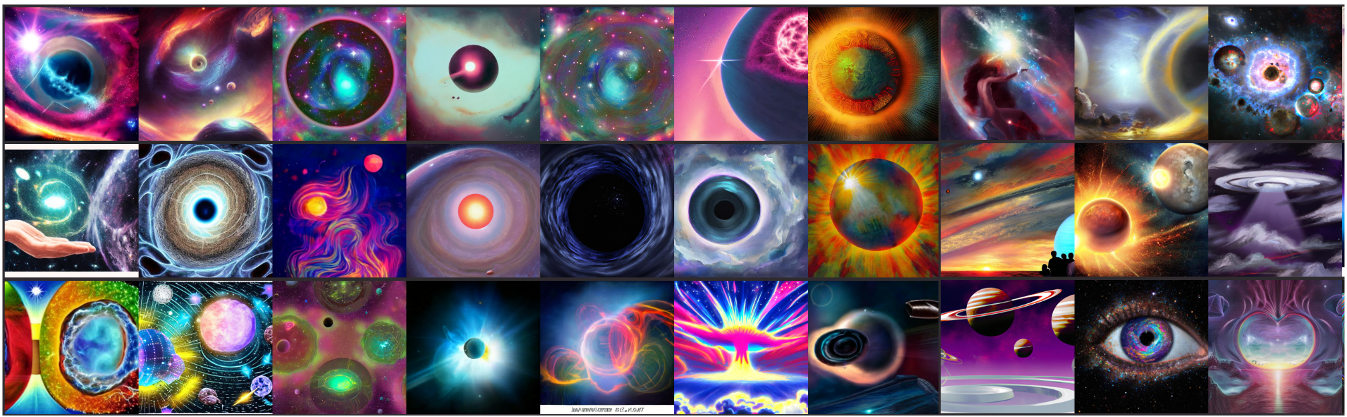}
        \caption{\myquote{You are a \textbf{graphic designer} with experience in visual clarity and technical image quality [...]}}
        \label{fig:big-qualitative-2-2}
    \end{subfigure}

    \begin{subfigure}[b]{0.80\textwidth}
        \centering
        \includegraphics[width=\textwidth]{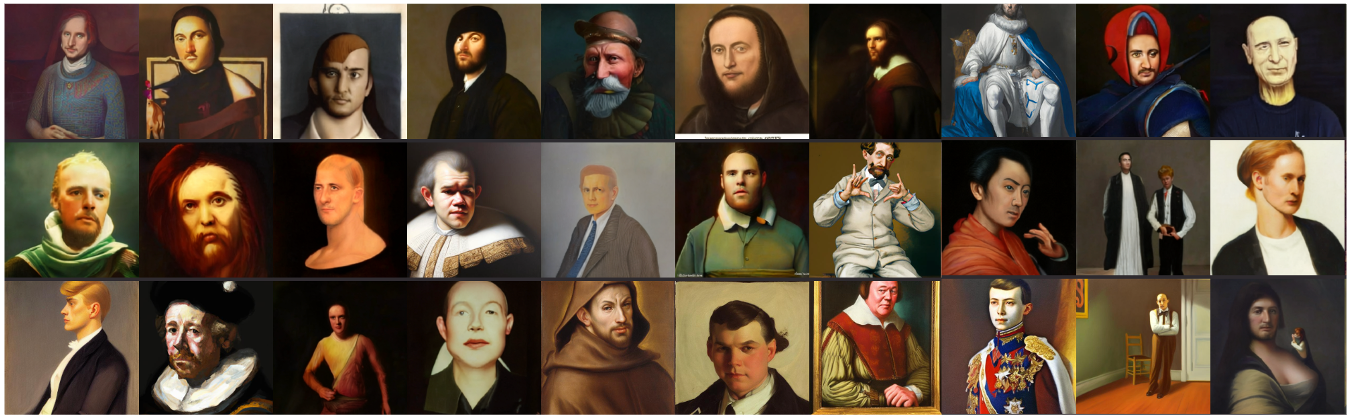}
        \caption{\myquote{You are an \textbf{art historian} with a keen eye for visual composition and color balance [...]}}
        \label{fig:big-qualitative-2-3}
    \end{subfigure}
    
    \begin{subfigure}[b]{0.80\textwidth}
        \centering
        \includegraphics[width=\textwidth]{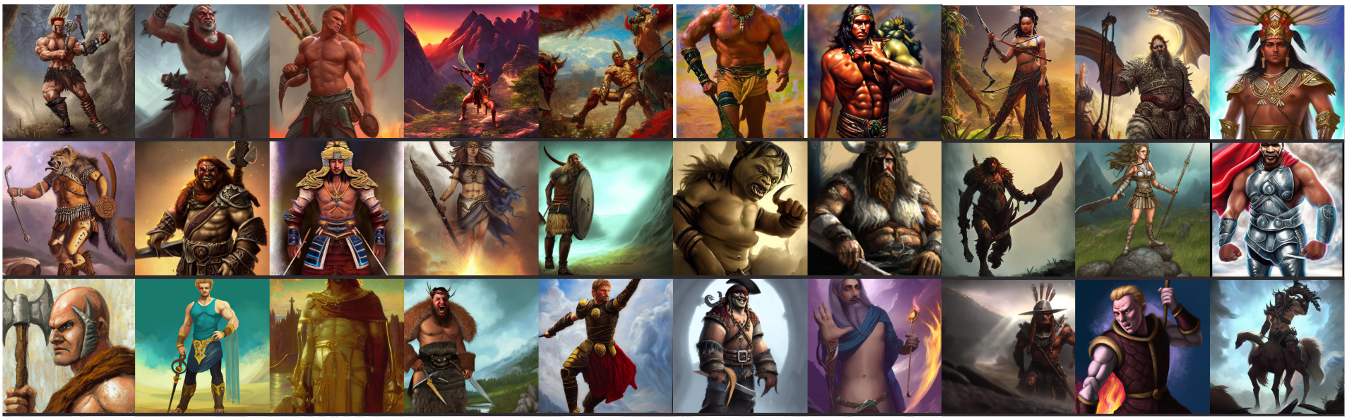}
        \caption{\myquote{You are an \textbf{AI ethics specialist} focusing on ensuring accurate and unbiased image representations [...]}}
        \label{fig:big-qualitative-2-4}
    \end{subfigure}
    
    \caption{Image clusters and their corresponding highest weighted prompts (or lack thereof) when using $K{=}64$, $N{=}200$.}
    \label{fig:big-qualitative-2}
\end{figure*}

\clearpage 

\section{MJBench Synthetic Preference Examples}
\label{sec:supp-synthetic}
\begin{figure*}[!h]
    \centering
    \begin{subfigure}[b]{0.80\textwidth}
        \centering
        \includegraphics[width=\textwidth]{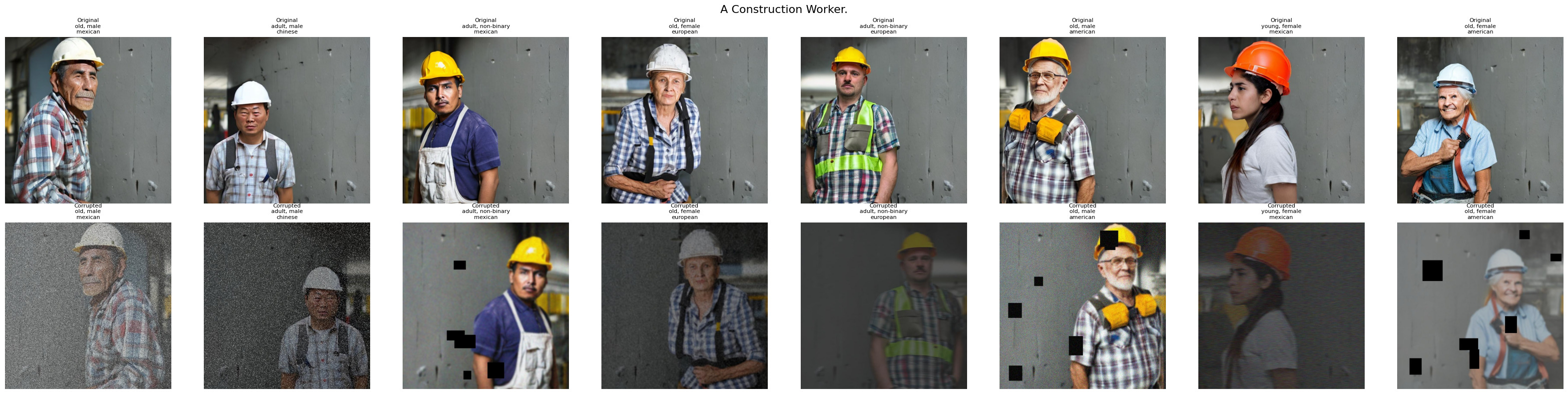}
        \label{fig:big-synthetic-3}
    \end{subfigure}
    
    \begin{subfigure}[b]{0.80\textwidth}
        \centering
        \includegraphics[width=\textwidth]{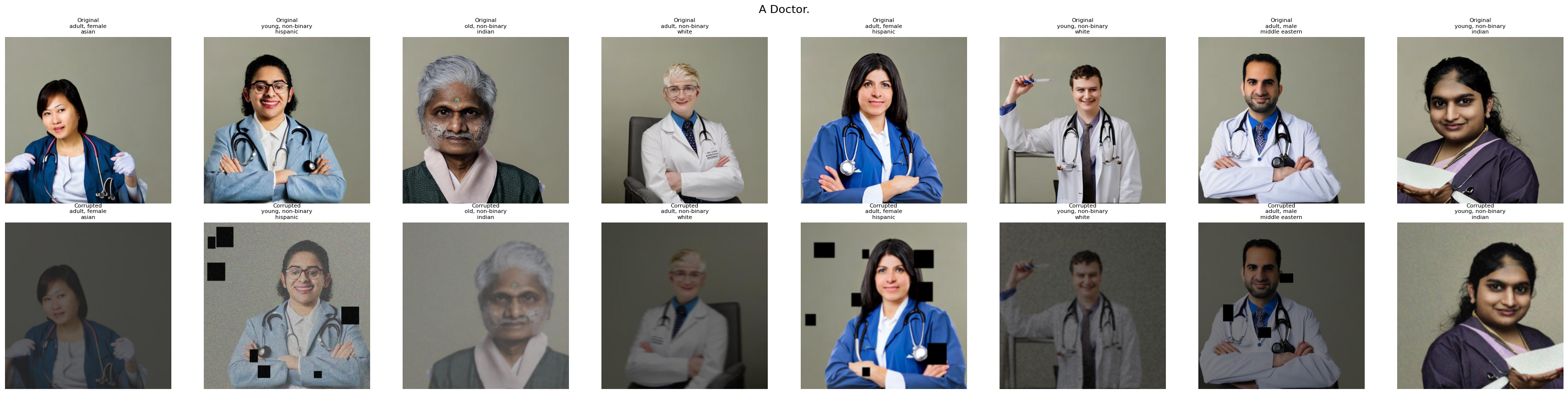}
        \label{fig:big-synthetic-4}
    \end{subfigure}

    \begin{subfigure}[b]{0.80\textwidth}
        \centering
        \includegraphics[width=\textwidth]{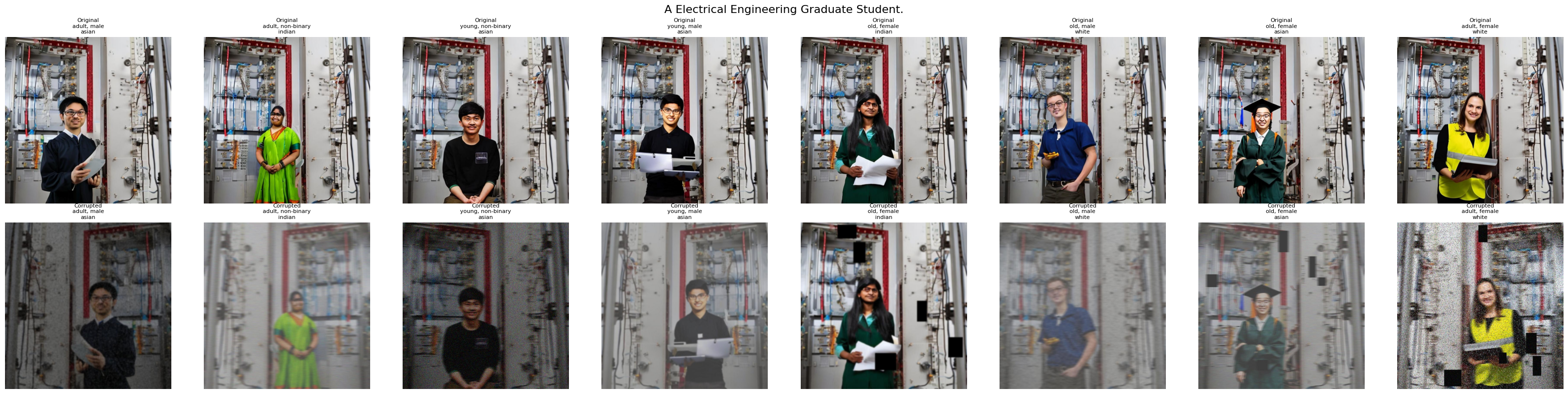}
        \label{fig:big-synthetic-5}
    \end{subfigure}

    \begin{subfigure}[b]{0.80\textwidth}
        \centering
        \includegraphics[width=\textwidth]{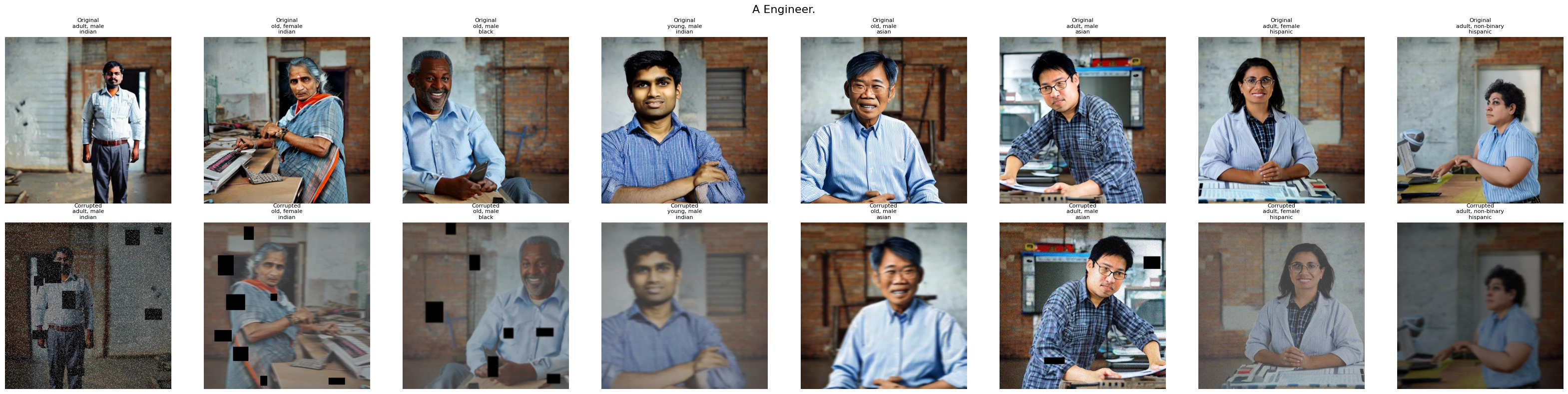}
        \label{fig:big-synthetic-6}
    \end{subfigure}

    \begin{subfigure}[b]{0.80\textwidth}
        \centering
        \includegraphics[width=\textwidth]{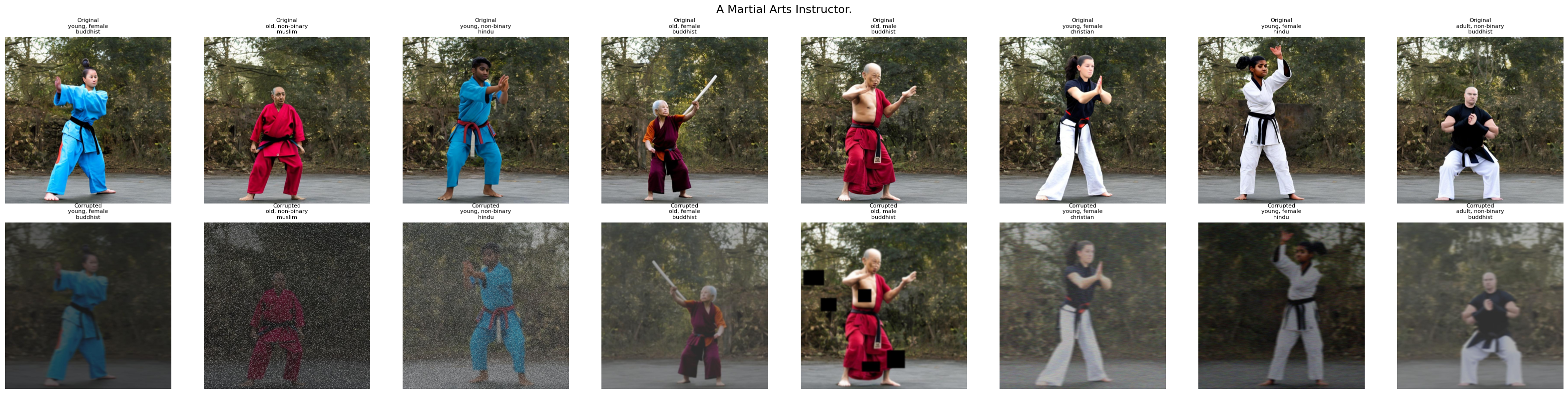}
        \label{fig:big-synthetic-7}
    \end{subfigure}
    \caption{Synthetic images preference pairs generated from MJBench-Bias.}
    \label{fig:big-synthetic}
\end{figure*}

\end{document}